\documentclass[letterpaper]{article}
\usepackage{aaai2026}
\usepackage{mathtools}
\usepackage{times}
\usepackage{helvet}
\usepackage{courier}
\usepackage[hyphens]{url}
\usepackage{graphicx}
\usepackage{natbib}
\usepackage{caption}
\usepackage{algorithm}
\usepackage{algpseudocode}  
\usepackage{amsmath}  
\usepackage{amssymb}
\usepackage{etoc}
\usepackage{xcolor} 
\usepackage{listings}
\usepackage{threeparttable}
\usepackage{newfloat}
\usepackage{listings}
\usepackage{multirow}
\usepackage{subcaption}
\usepackage{booktabs}
\usepackage[table,xcdraw]{xcolor}
\usepackage{tabularx}
\usepackage[toc,page]{appendix}
\DeclareCaptionStyle{ruled}{labelfont=normalfont,labelsep=colon,strut=off} 
\floatstyle{ruled}
\newfloat{listing}{tb}{lst}{}
\floatname{listing}{Listing}
\title{EAGLE: Episodic Appearance- and Geometry-aware Memory for Unified 2D-3D Visual Query Localization in Egocentric Vision}
\author{
    Yifei Cao\textsuperscript{\rm 1},
    Yu Liu\textsuperscript{\rm 1}\equalcontrib,
    Guolong Wang\textsuperscript{\rm 2}\equalcontrib,\\
    Zhu Liu\textsuperscript{\rm 1},
    Kai Wang\textsuperscript{\rm 3}, Xianjie Zhang\textsuperscript{\rm 4}, Jizhe Yu\textsuperscript{\rm 1}, Xun Tu\textsuperscript{\rm 1}
}
\affiliations{
    \textsuperscript{\rm 1}Dalian University of Technology,
    \textsuperscript{\rm 2}University of International Business and Economics\\
    \textsuperscript{\rm 3}Harbin Institute of Technology,
    \textsuperscript{\rm 4}Academy of Satellite Application Innovation,CASC\\
    \{yfcao,liuzhu,yujizhe,tuxun\}@mail.dlut.edu.cn, yuliu@dlut.edu.cn, wangguolong@uibe.edu.cn, kai\_wang@hit.edu.cn,xj\_zh@foxmail.com
}

\usepackage{bibentry}

\begin{document}

\maketitle
\begin{figure*}[tp]
\centering
\includegraphics[width=\textwidth,height=0.8\columnwidth]{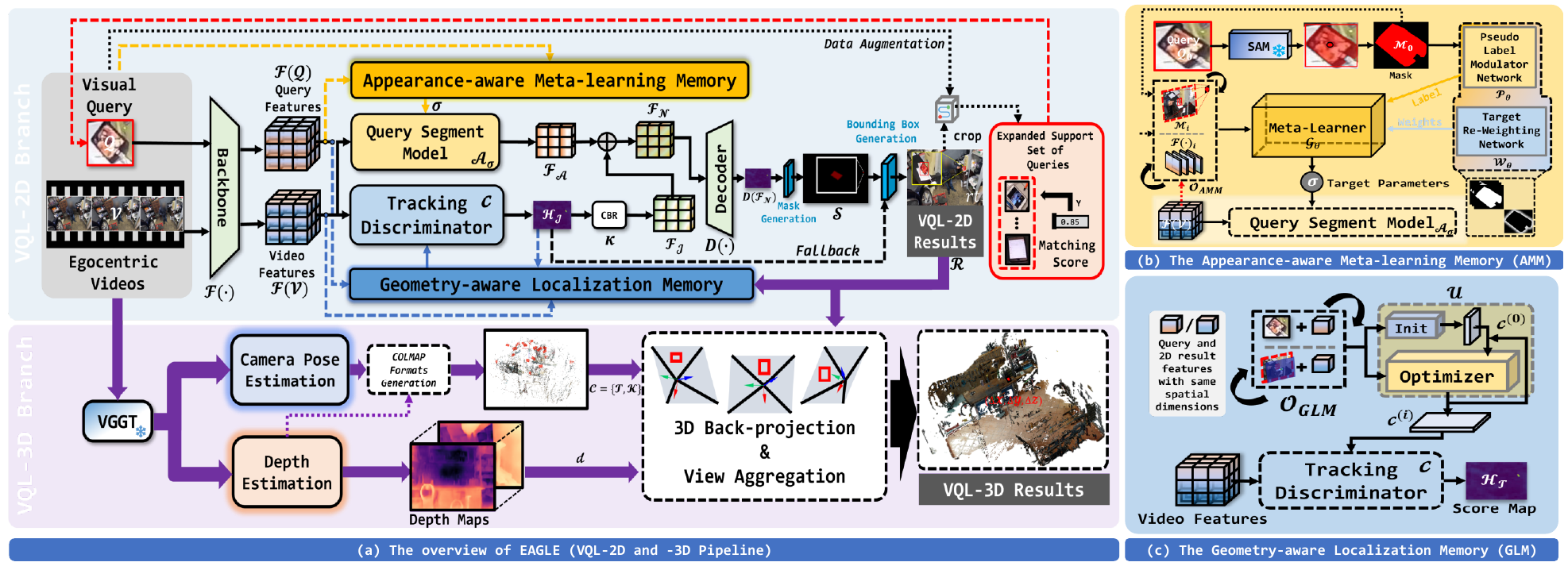}
\caption{\textbf{Overview of EAGLE}. Our framework consists of two branches: VQL-2D and -3D. The VQL-2D features a dual-branch architecture built upon a shared backbone. The segmentation branch serves as a precise identifier, guided by an appearance-aware meta-learning memory (b) to generate pixel-level masks with fine-grained semantic cues. The tracking branch, acting as a navigator, is driven by a geometry-aware localization memory (c) to produce a discriminative score map robust to egocentric view changes. Finally, a decoder fuses the outputs from both branches to yield the results. 3D branch leverages the VGGT to jointly process the 2D results, camera pose, and depth, ultimately predicting the positional offset of the query in 3D space.}
\label{fig:overall}
\end{figure*}
\begin{abstract}
Egocentric visual query localization is vital for embodied AI and VR/AR, yet remains challenging due to camera motion, viewpoint changes, and appearance variations. We present \textbf{EAGLE}, a novel framework that leverages \textbf{e}pisodic \textbf{a}ppearance- and \textbf{g}eometry-aware memory to achieve unified 2D-3D visual query \textbf{l}ocalization in \textbf{e}gocentric vision. Inspired by avian memory consolidation, EAGLE synergistically integrates segmentation guided by an appearance-aware meta-learning memory (AMM), with tracking driven by a geometry-aware localization memory (GLM). This memory consolidation mechanism, through structured appearance and geometry memory banks, stores high-confidence retrieval samples, effectively supporting both long- and short-term modeling of target appearance variations. This enables precise contour delineation with robust spatial discrimination, leading to significantly improved retrieval accuracy. Furthermore, by integrating the VQL-2D output with a visual geometry grounded Transformer (VGGT), we achieve a efficient unification of 2D and 3D tasks, enabling rapid and accurate back-projection into 3D space. Our method achieves state-of-the-art performance on the Ego4D-VQ benchmark.
\end{abstract}

\section{Introduction}
Visual query localization (VQL) is a fundamental task in egocentric episodic memory, which aims to spatio-temporally localize the final occurrence of a target within a video, guided by a visual crop. This capability serves as a cornerstone for downstream applications such as virtuality or augmented reality (VR/AR), embodied AI, etc..\cite{ego4d,egooutlook,vqloc,egoloc,wallet,egotracks,ego3dt} Nevertheless, the egocentric perspective presents challenges, including drastic camera motion, severe motion blur, and variations in object appearance and scale. These factors frequently lead to retrieval failures, critically impeding progress in this field.

We revisit the prevalent “detect-then-track” pipeline for VQL and expose its limitations. (i) The aforementioned paradigm couples a detector (the \textit{identifier}) with a tracker (the \textit{navigator}), manifests following principal shortcomings: the identifier's reliance on bounding boxes leads to the inclusion of substantial background pixels, especially for non-rigid targets. These view-dependent background signals can contaminate the target's appearance, thus impairing search accuracy. Concurrently, the navigator lacks the robustness to handle challenges such as extreme view changes, drastic scale transformations, motion blur, and similar distractors. (ii) During retrieval, relying on a static, low-visibility query is often insufficient to capture the target’s appearance variations over time; instead, humans typically integrate multiple visual cues from different temporal snapshots to compensate for the limitations of a single-frame query. (iii) The prevailing VQL paradigms have not yet achieved a natural unification between 2D and 3D tasks, notwithstanding the intimate correlation between them in real world. 

The avian visual system offers profound insights into this problem. Eagle, for instance, exhibit remarkable episodic memory and spatial localization, enabling them to retain the appearance of a specific object over extended periods and precisely recall its spatiotemporal position\cite{episodicmemory,episodic}. This capability originates from a “memory consolidation” mechanism: initially, the system rapidly forms a short-term memory “imprint” of key features. Subsequently, through continuous observation, it actively disambiguates the target from its evolving environment, eventually solidifying this information into a stable, long-term memory. Inspired by this biological schema, an ideal VQL system should transcend the passive storage of static visual cues. It must instead actively filter and encode object that is both critical for long-term recognition and inherently stable. Such a selective memorization mechanism enhances the model's robustness to environmental distractors, thereby facilitating precise target retrieval within dynamic and variable egocentric videos. 

To address the aforementioned challenges, we introduce a novel VQL framework that integrates two parallel branches—segmentation and tracking—which serve as the \textit{identifier} and \textit{navigator}, respectively. Inspired by the memory consolidation mechanism in avian vision systems, both branches are driven by independent online episodic memory banks. These banks dynamically update their support sets by continuously filtering for high-confidence visual observations, enabling long-term adaptive learning of the target's state. Figure.\ref{fig:overall} (a) illustrates the framework. Specifically, the segmentation branch incorporates an appearance-aware meta-learning memory (AMM). Starting with an initial mask generated by SAM\cite{sam}, it constructs rich supervisory signals via a pseudo-label modulator network and a target re-weighting network. This module populates its episodic memory bank with high-confidence retrieval results and employs the steepest descent method to rapidly update model parameters, facilitating continuous meta-learning of the target's appearance for pixel-level accurate retrieval. The tracking branch employs a discriminative correlation filter (DCF)\cite{dcf} to rectify potential instance misassignments from the segmentation branch. By maintaining a memory bank that stores both the static initial query and dynamic high-confidence observations, this branch proactively encodes stable geometric information crucial for long-term recognition. This provides strong constraints on the segmentation results and enables robust, long-term modeling of target appearance and scale variations. Furthermore, to efficiently unify 2D and 3D tasks, we feed the 2D-VQL output into a VGGT\cite{vggt} to refine camera pose and depth estimations. Subsequently, the 2D segmentation results are back-projected into 3D space, yielding a unified 3D-VQL output. This design significantly enhances inference efficiency and memory utilization, laying a solid foundation for more realistic embodied episodic memory retrieval. In summary, the contributions are as follow:
\begin{itemize}
    \item We propose a novel dual-branch framework for egocentric VQL that synergizes segmentation and tracking to leverage precise, pixel-level appearance cues and robust geometric-temporal constraints, overcoming the limitations of traditional “detect-then-track” paradigms.
    \item Inspired by biological cognitive processes, we devise an online episodic memory bank-driven memory consolidation mechanism for both branches. This mechanism selectively incorporates high-confidence observations into structured appearance and geometry memory banks, enabling continuous, long-term adaptation to target variations while mitigating interference from distractors.
    \item We achieve an efficient and unified 2D-to-3D localization by back-projecting the refined 2D VQL outputs into 3D space, significantly enhancing the applicability of our method for embodied AI scenarios.
    \item Comprehensive experiments on the Ego4D-VQ benchmark validate the superiority of our method, demonstrating state-of-the-art performance in accuracy, robustness, and efficiency.
\end{itemize}
\section{Related Work}
\textbf{Few-shot Visual Object Tracking}.
VQL is fundamentally a few-shot, open-world tracking problem\cite{vq2d,prvql,relocate,egotracks,it3dego}. While meta-learning-based trackers\cite{d3s,dimp,metaupdater,metatracker,LTMU,kys,metarcnn} show promise due to their rapid adaptation capabilities, they primarily address “how to update”. We argue that the core challenge in VQL is “what to learn”—constructing an optimal and robust target representation. To this end, inspired by biological memory consolidation, we introduce a dual-memory mechanism. Our model jointly leverages segmentation and tracking to dynamically build online memory banks from query-response interactions within the video, enabling it to actively distill discriminative features and robustly handle the target's evolving states.

\noindent\textbf{2D \& 3D Visual Query Localization.} VQL, comprising both 2D and 3D tasks, aims to spatio-temporally localize a target's final occurrence. Early methods for VQL-2D relied on multi-stage “detect-then-track” pipelines \cite{ego4d, wallet,vq2d}, while more recent works like PRVQL \cite{prvql}, HERO-VQL\cite{herovql}, and RELOCATE \cite{relocate} have streamlined this into single-stage or training-free frameworks. For VQL-3D, existing approaches \cite{egocol, egoloc,egolocv1} typically depend on Structure-from-Motion (SfM)\cite{sfm} techniques like COLMAP\cite{colmap} for pose estimation. However, SfM is often fragile, failing in texture-less or high-motion scenarios, and leads to fragmented, inefficient pipelines. To overcome these limitations, we first address VQL-2D by integrating memory-guided segmentation and tracking for robust retrieval. We then tackle VQL-3D by replacing the brittle SfM-based pose estimation with VGGT \cite{vggt}. Compared to other learning-based geometric foundation models\cite{nerf,neuraldiff,3dgs,dust3r}, VGGT more efficiently infers camera pose and depth in a feedforward pass. This substitution not only enhances 3D displacement accuracy but also unifies the 2D and 3D tasks into a single, efficient pipeline.
\section{The Proposed Method}
\subsection{The AMM-guided segmentation branch}
As shown in Figure.\ref{fig:overall}(b), we leverage SAM to obtain an initial segmentation mask, denoted as $\mathcal{M}_0 = SAM(\mathcal{Q}_0)$, for the visual query $\mathcal{Q}_0$. To provide richer supervision for the subsequent meta-learning process, we introduce a pseudo-label modulator network, $\mathcal{P}_\theta$. This trainable, lightweight convolutional network transforms the input binary mask into a multi-channel pseudo-label, $\mathcal{QM}_i = \mathcal{P}_\theta(\mathcal{M}_i)$, which encapsulates rich semantic information such as boundaries and centers. Concurrently, we design a target re-weighting network ($\mathcal{W}_\theta$) with a similar architecture to $\mathcal{P}_\theta$, aiming to guide the loss function to focus on critical regions of the target. To facilitate an efficient and differentiable meta-learning process, $\mathcal{A}_\sigma$ maps a tensor from $\mathbb{R}^{H\times W\times C}$ to $\mathbb{R}^{H\times W\times D}$. This is formulated as $\mathcal{A}_\sigma(x) = x * \sigma$, where $\sigma$ represents the weights of a $\mathbb{R}^{K\times K\times C\times D}$ convolutional layer, and $*$ denotes the convolution operation. Our proposed meta-learner, $\mathcal{G}_\theta$, optimizes the parameters $\sigma$ by minimizing the weighted squared error between the prediction of the query segment model $\mathcal{A}_\sigma$ and the corresponding pseudo-labels over the episodic memory bank $\mathcal{O}_{AMM}$. Initially, $\mathcal{O}_{AMM}$ contains only the query sample $(\mathcal{F}_{\mathcal{Q}_0}, \mathcal{M}_0)$. Subsequently, it is updated by incorporating segmentation retrieval results extracted from the video stream. Specifically, a segmentation sample based on the segmentation result $\mathcal{S}_{I_i}$ generated by the model on a retrieved video frame $I_i$ is added to the memory bank only if the target exhibits low entropy or high certainty in its corresponding confidence map. This condition is met when the mean confidence score within the predicted mask region exceeds a threshold, $s_{conf}>=0.6$. The corresponding region's segmentation sample is then cropped and resized proportionally ($\mathcal{M}_i = \text{crop}(\mathcal{S}_{I_i})$) before being added to the memory bank, updating it as $\mathcal{O}_{AMM} = \mathcal{O}_{AMM} \cup \{(\mathcal{F}(I_i), \mathcal{M}_i)\}$. The loss function is then defined as:
\begin{equation} \label{equ:lossseg}
\mathcal{L}(\sigma) = \frac{1}{2}\sum||\mathcal{W}_\theta(\mathcal{M}_i) \odot (\mathcal{A}_\sigma(\mathcal{F}_i) - \mathcal{QM}_i)||^2 + \frac{\delta}{2}||\sigma||^2,
\end{equation}
where $\delta$ is a learnable regularizer. $(\mathcal{F}_i, \mathcal{M}_i)$ is the $i$-th feature-pseudo-label pair from the memory $\mathcal{O}_{AMM}$. Given that Eq.(\ref{equ:lossseg}) defines a convex quadratic objective with respect to $\sigma$, admitting a closed-form solution in either its primal or dual form, we solve it using the iterative steepest descent method. At each iteration, given the current estimate $\sigma_i$, the step size $\alpha^i$ is chosen to minimize the loss along the gradient direction, i.e., $\alpha^i = \arg\min_{\alpha} \mathcal{L}(\sigma_i - \alpha g^i)$, where $g^i = \nabla \mathcal{L}(\sigma_i)$ represents the gradient of the loss function evaluated at $\sigma_i$. The parameters are iteratively updated using the steepest descent method, as below:
\begin{equation}
\label{equ:steepupdate}
\begin{aligned}
    \sigma_{i+1} &= \sigma_i - \alpha^i g^i, \\
    \alpha^i &= \frac{\|g^i\|^2}{\sum_{i} \|\mathcal{W}_{\theta}(\mathcal{M}_i) \odot (\mathcal{F}_i * g^i)\|^2 + \delta\|g^i\|^2}, \\
    g^i &= \sum_{i} \mathcal{F}_i *^T \left( \mathcal{W}_{\theta}^2(\mathcal{M}_i) \odot (\mathcal{F}_i * \sigma_i - \mathcal{P}_\theta(\mathcal{M}_i)) \right) + \delta\sigma_i.
\end{aligned}
\end{equation}
where $*^T$ represents the transposed convolution. After $i$ iterations, the resulting parameter $\sigma_i$ is differentiable with respect to all network parameters $\theta$. The function $\mathcal{G}_\theta(\mathcal{O}_{AMM}, \sigma_0) = \sigma_N$ is defined by performing $N$ iterations of the steepest descent update in Eq.(\ref{equ:steepupdate}), initialized with $\sigma_0$. Leveraging the rapid convergence of the steepest descent method, our optimization-based paradigm facilitates efficient updates to $\sigma$. New samples are added to $\mathcal{O}_{AMM}$, and a few iterations are performed starting from the current estimate of $\sigma$ (Eq.(\ref{equ:steepupdate})). 
\subsection{The GLM-guided tracking branch}
To mitigate potential instance misassignments from the segmentation branch, we introduce a tracking-based navigator. This branch employs DCF to provide robust geometric constraints on the segmentation output. We construct a geometric localization memory (GLM) bank, $\mathcal{O}_{GLM}$, which stores both the initial static query and a dynamic set of high-confidence historical observations retrieved by the 2D branch. Figure.\ref{fig:overall}(c) illustrates the architecture of the GLM. The initial query features and their corresponding Gaussian labels are stored as static snapshots in the episodic memory bank, $\mathcal{O}_{GLM}$. Concurrently, feature maps corresponding to the 2D retrieved target regions from the query-expanded support set are isotropically scaled by a factor of 1.5. These scaled features, along with their Gaussian labels, are then added to $\mathcal{O}_{GLM}$ as dynamic snapshots, which are updated following a FIFO policy. The model updater, $\mathcal{U}$, serves as the core module for constructing the DCF. It takes $\mathcal{O}_{GLM} = \{(\mathcal{F}(\cdot)_i, G(\cdot)_i)\}_{i=1}^n$ as input, where $\mathcal{F}(\cdot)_i$ represents either the query or the result of 2D branch features, and $G(\cdot)_i$ is the corresponding Gaussian label. The objective is to learn a set of convolutional kernel weights, $c$, to construct the target model, formulated as $c = \mathcal{U}(\mathcal{O}_{GLM})$.

We employ a least-squares regression loss function to supervise the training of the DCF, defined as follows:
\begin{equation}
\label{equ:1}
   \mathcal{L}(c) = \frac{1}{|\mathcal{O}_{GLM}|} \sum_{(\mathcal{F}(\cdot),G(\cdot))\in\mathcal{O}_{GLM}} \|\mathcal{H}(\mathcal{H}_\mathcal{J}, G)\|^2 + \|\lambda c\|^2. 
\end{equation}
Here, $\lambda$ is the regularization coefficient. $\mathcal{H}(\mathcal{H}_\mathcal{J}, G)$ represents the spatial residual between the predicted score $\mathcal{H}_\mathcal{J} = \mathcal{F}(\cdot) * c$ and the corresponding Gaussian label $G$. Drawing inspiration from the efficacy of Hinge loss in addressing data imbalance, we formulate the residual as:
\begin{equation}
\label{equ:2}
    \mathcal{H} = sw_G \cdot (\mathcal{S}_i \mathcal{H}_\mathcal{J} + (1-\mathcal{S}_i) \max(0, \mathcal{H}_\mathcal{J}) - G_i).
\end{equation}
Here, $sw_G$ denotes a spatial weighting function contingent on the Gaussian label $G$, which assigns higher weights to positions proximate to the target's center and lower weights to those in the background. $\mathcal{S}_i$ represents the corresponding distinct binary masks, originating from the initial query only when $i=0$, and otherwise from 2D retrieval results. This formulation precisely fits the target score $G_i$ when $\mathcal{S}_i=1$; employs the hinge branch when $\mathcal{S}_i=0$, focusing solely on whether $\mathcal{H}_\mathcal{J} > 0$; and automatically learns the boundary distance of the target object when $0 < \mathcal{S}_i < 1$. This approach enables the loss function to smoothly transition between least squares and hinge based on the pixel's distance to the target center.

Previous works optimize DCF through a limited number of iteration steps\cite{atom,kys}:
$c^{(i+1)} = c^{(i)} - \beta \nabla \mathcal{L}(c^{(i)})$. We adopt a more refined optimization strategy—steepest descent iteration—to compute an adaptive step size. First, we perform a quadratic approximation at the current estimate $c^{(i)}$:
\begin{equation}
\label{equ:3}
\begin{split}
    \mathcal{L}(c) \approx \hat{\mathcal{L}}(c) ={}& \frac{1}{2} (c - c^{(i)})^\top PDS^{(i)} (c - c^{(i)}) \\
    & + (c - c^{(i)})^\top \nabla \mathcal{L}(c^{(i)}) + \mathcal{L}(c^{(i)}).
\end{split}
\end{equation}
Both $c$ and $c^{(i)}$ are treated as vectors, and $PDS^{(i)}$ is a positive definite matrix. Subsequently, in the gradient direction (Eq.(\ref{equ:3})), we solve for the step size $\beta$ that minimizes the approximate loss by setting $\frac{\partial}{\partial \beta} \hat{\mathcal{L}} (c^{(i)} - \beta \nabla \mathcal{L}(c^{(i)})) = 0$:
\begin{equation}
\label{equ:4}
    \beta = \frac{\nabla \mathcal{L}(c^{(i)})^\top \nabla \mathcal{L}(c^{(i)})}{\nabla \mathcal{L}(c^{(i)})^\top PDS^{(i)} \nabla \mathcal{L}(c^{(i)})}.
\end{equation}
This formula computes the step size $\beta$ in the update Eq.(\ref{equ:3}). Similar to \cite{atom,dimp}, we set $PDS^{(i)} = \frac{\partial^2 \mathcal{L}}{\partial c^2}(c^{(i)})$, which is the Hessian matrix of the loss function (Eq.(\ref{equ:1})), employing the second-order Taylor expansion from Eq.(\ref{equ:4}). For Eq.(\ref{equ:1}), the Gauss-Newton method based on first-order derivatives is more efficient, where $PDS^{(i)} = (J^{(i)})^\top J^{(i)}$, and $J^{(i)}$ is the Jacobian matrix of the residuals at $c^{(i)}$. Beyond the initial phase, during online inference, if the score map $\mathcal{H}_\mathcal{J}$ fails to consistently produce high responses (exceeding 60\% of the historical frame length), we update the DCF using the original static snapshot; otherwise, we employ the most recent dynamic snapshot to update the model. This approach enables learning filter weights that balance static and dynamic scenarios, representing the responses of the query target against the background across varying times, scenes, and states via geometric response scores.
\begin{figure}[t]
\centering
\includegraphics[width=0.47\textwidth]{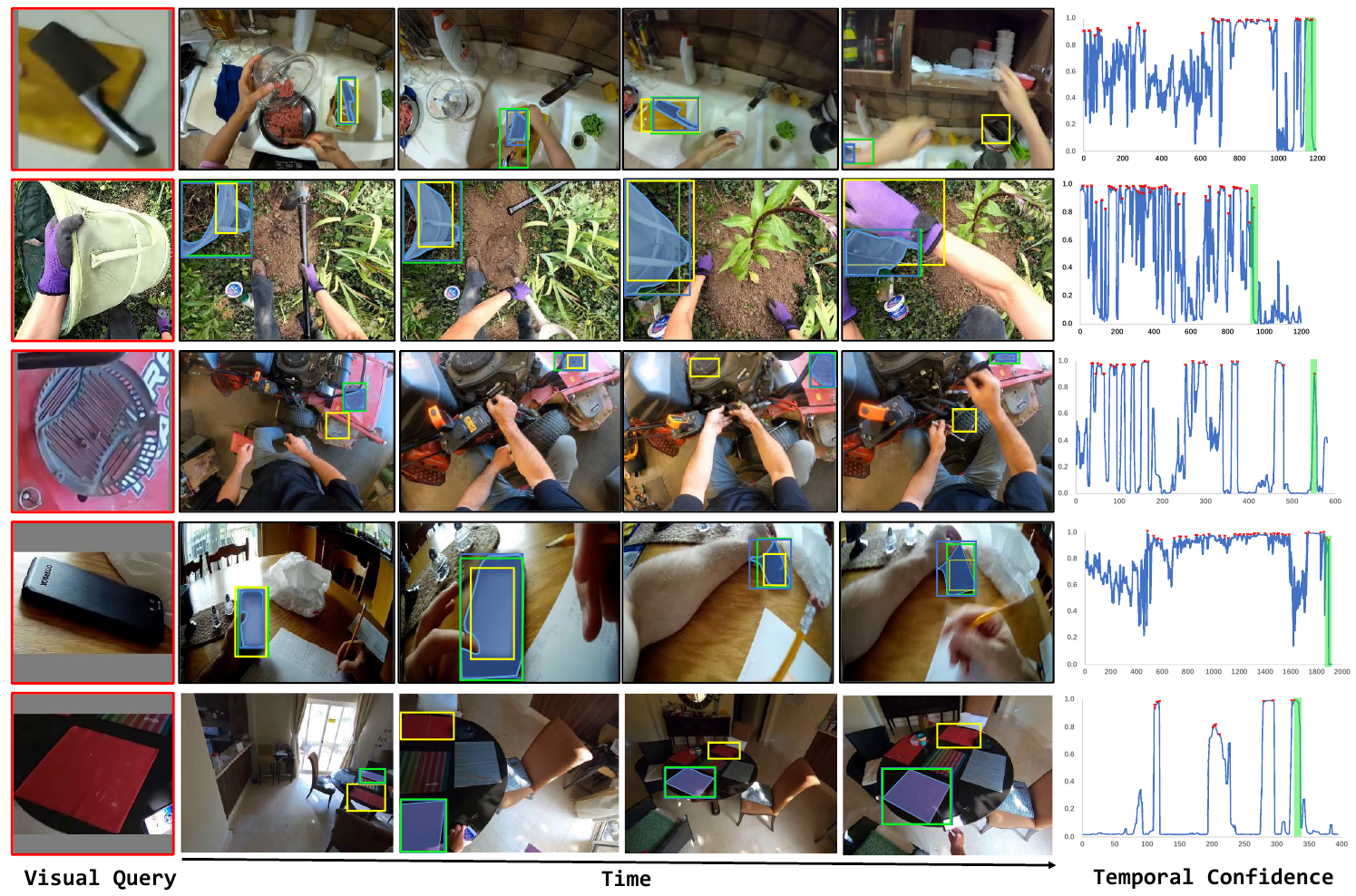}
\caption{\textbf{Visualization of qualitative results of VQL-2D.} Each row presents the visual query, video frames, the predicted trajectories from both EAGLE, VQLoc, and the ground truth. Additionally, the temporal confidence curve predicted by EAGLE is shown, with the green shaded region indicating the ground-truth interval.}
\label{fig:quality}
\end{figure}
\subsection{Dual Branches Integration}
To mitigate the risk of individual branch degradation, we adopt a dual-branch framework wherein each branch provides complementary prior knowledge. Specifically, the tracking branch's score map, $\mathcal{H}_\mathcal{J}$, is encoded by a conv-bn-relu block, $\kappa_\theta$, resulting in $\mathcal{F}_\mathcal{J}=\kappa_\theta (\mathcal{H}_\mathcal{J})$, which is dimensionality-matched to the mask encoding, $\mathcal{F}_\mathcal{A}$. These features are then fused via element-wise addition, yielding a combined feature representation, $\mathcal{F}_\mathcal{N}=\mathcal{F}_\mathcal{A}+\mathcal{F}_\mathcal{J}$. The decoder, $\mathbf{D}$, processes $\mathcal{F}_\mathcal{N}$ to produce a segmentation score map, $\mathbf{D}(\mathcal{F}_\mathcal{N})$. Finally, mask generation yields the segmentation result, $\mathcal{S}$, at the original resolution, and the bbox trajectory, $\mathcal{R}$, used for evaluation is derived from the minimum bounding rectangle enclosing the connected components of $\mathcal{S}$.
\subsection{VGGT-Powered 3D Visual Localization}
\label{sec:vggt3d}
Given an input video sequence $\mathcal{V}=\{I_i\}_{i=1}^{N}$ consisting of $N$ frames, VGGT jointly infers per-frame geometric cues in a single, end-to-end pass. For each image $I_i$, the model outputs the camera parameters $\mathcal{C}_i=\{\mathcal{T}_i,\,\mathcal{K}_i\}$, a dense depth map $\mathcal{D}_i$, and a pixel-wise depth uncertainty map $\mathcal{UD}_i$. This pipeline reduces the processing time from the several minutes or even hours required by COLMAP to only a few seconds, while delivering geometry estimations of comparable accuracy. The resulting geometric priors form the backbone of the subsequent aggregation strategy. VGGT predicts 3D geometry in a self-consistent, yet arbitrary, coordinate system. To evaluate on the VQL-3D benchmark, we must align our predictions with the ground-truth matterport scan coordinate system for each sequence. This is achieved by solving for a 7-DoF similarity transformation, $\mathbf{T}_\eta \in \text{Sim}(3)$, which maps our predicted point cloud ($\mathbf{pc}^{\text{vggt}}$) to the ground-truth ($\mathbf{pc}^{\text{ms}}$). We formulate this as a least-squares optimization problem:
$\mathbf{T}_\eta = \underset{\mathbf{T} \in \text{Sim}(3)}{\arg\min} \sum_{j} || \mathbf{T} \cdot \mathbf{pc}_{j}^{\text{vggt}} - \mathbf{pc}_{j}^{\text{ms}} ||^2$,
where $j$ indexes corresponding 3D point pairs. Once the optimal transformation $\mathbf{T}_\eta$ is found, we apply it to all predicted camera poses and 3D structures, thereby unifying them into the canonical benchmark coordinate system for evaluation.

\begin{table}[tbp]
\resizebox{\columnwidth}{!}{%
\begin{threeparttable}
\begin{tabular}{ccccccccc}
\hline
 &
  \multicolumn{4}{c}{\textbf{VQ2D Test Server Leaderboard}} &
  \multicolumn{4}{c}{\textbf{VQ2D Validation Set}} \\ \cline{2-9} 
\multirow{-2}{*}{Method} &
  \textbf{tAP}$_{25}$ $\uparrow$ &
  \textbf{stAP}$_{25}$ $\uparrow$ &
  \textbf{Rec.(\%)} $\uparrow$ &
  \textbf{Succ.(\%)} $\uparrow$ &
  \textbf{tAP}$_{25}$ $\uparrow$ &
  \textbf{stAP}$_{25}$ $\uparrow$ &
  \textbf{Rec.(\%)} $\uparrow$ &
  \textbf{Succ.(\%)} $\uparrow$ \\ \hline
Ego4D baseline [CVPR'22] &
  0.20 &
  0.13 &
  32.20 &
  39.80 &
  0.22 &
  0.15 &
  32.92 &
  43.24 \\
NFM [Ego4D 2022 Winner] &
  0.24 &
  0.17 &
  35.29 &
  43.07 &
  0.26 &
  0.19 &
  37.88 &
  47.90 \\
CocoFormer [CVPR'23] &
  0.25 &
  0.18 &
  42.34 &
  48.37 &
  0.26 &
  0.19 &
  37.67 &
  47.68 \\
STARK-S50 [ICCV'21] &
  - &
  - &
  - &
  - &
  0.08 &
  0.03 &
  11.35 &
  15.08 \\
STARK-S50($^\dagger$) &
  - &
  - &
  - &
  - &
  0.29 &
  0.20 &
  35.57 &
  45.20 \\
STARK-S101 &
  - &
  - &
  - &
  - &
  0.10 &
  0.04 &
  12.41 &
  18.70 \\
STARK-S101($^\dagger$) &
  - &
  - &
  - &
  - &
  0.30 &
  0.21 &
  41.11 &
  48.03 \\
VQLoc [NeurIPS'23] &
  0.32 &
  0.24 &
  45.10 &
  55.88 &
  0.31 &
  0.22 &
  47.05 &
  55.89 \\
HERO-VQL [BMVC'25] &
  0.37 &
  0.28 &
  45.32 &
  60.72 &
  0.38 &
  0.28 &
  44.90 &
  61.10 \\
PRVQL [ICCV'25] &
  0.37 &
  0.28 &
  45.70 &
  59.43 &
  0.35 &
  0.27 &
  47.87 &
  57.93 \\
RELOCATE [CVPR'25] &
  0.43 &
  0.35 &
  50.60 &
  60.10 &
  0.41 &
  0.33 &
  50.50 &
  58.03 \\
\rowcolor[HTML]{E0E0E0}
\textbf{EAGLE (Ours)} &
  \textbf{0.46} &
  \textbf{0.40} &
  \textbf{53.51} &
  \textbf{62.70} &
  \textbf{0.47} &
  \textbf{0.42} &
  \textbf{52.09} &
  \textbf{61.29} \\ \hline
\end{tabular}%
\end{threeparttable}
}
\caption{\textbf{Comparison results on Ego4D-VQ2D benchmark}. Ego4D provides the complete definitions of the evaluation metrics. $\dagger$ denotes the approach fine-tuned on EgoTracks, the same applies hereinafter.}
\label{tab:vq2d}
\end{table}
\begin{table}[]
\resizebox{\columnwidth}{!}{%
\begin{threeparttable}
\begin{tabular}{@{}c|ccccc|ccccc@{}}
\toprule
 &
  \multicolumn{5}{c|}{\textbf{VQ3D Test Server Leaderboard}} &
  \multicolumn{5}{c}{\textbf{VQ3D Validation Set}} \\ \cmidrule(l){2-11} 
\multirow{-2}{*}{\textbf{Method}} &
  \textbf{Succ.(\%)} $\uparrow$ &
  \textbf{Succ.*(\%)} $\uparrow$ &
  \textbf{L2} $\downarrow$ &
  \textbf{Angle} $\downarrow$ &
  \textbf{QwP(\%)} $\uparrow$ &
  \textbf{Succ.(\%)} $\uparrow$ &
  \textbf{Succ.*(\%)} $\uparrow$ &
  \textbf{L2} $\downarrow$ &
  \textbf{Angle} $\downarrow$ &
  \textbf{QwP(\%)} $\uparrow$ \\ \midrule
Ego4D[CVPR'22] &
  7.95 &
  48.61 &
  4.64 &
  1.31 &
  0.16 &
  - &
  - &
  - &
  - &
  - \\
Ego4D*[CVPR'22] &
  8.71 &
  51.47 &
  4.93 &
  1.23 &
  15.15 &
  1.22 &
  30.77 &
  5.98 &
  1.60 &
  1.83 \\
Eivul[Ego4D 2022 Challenger] &
  25.76 &
  38.74 &
  8.97 &
  1.21 &
  66.29 &
  73.78 &
  91.45 &
  2.05 &
  0.82 &
  80.49 \\
CocoFormer[CVPR'23] &
  9.09 &
  50.60 &
  4.23 &
  1.23 &
  16.29 &
  - &
  - &
  - &
  - &
  - \\
EgoCOL[Ego4D 2023 Winner] &
  62.88 &
  85.27 &
  2.37 &
  0.53 &
  74.62 &
  59.15 &
  93.39 &
  2.31 &
  0.58 &
  63.42 \\
EgoLoc[ICCV'23] &
  87.12 &
  96.14 &
  1.86 &
  0.92 &
  90.53 &
  80.49 &
  98.14 &
  1.45 &
  0.61 &
  82.32 \\
EgoLoc-v1[CVPR'24] &
  88.64 &
  96.15 &
  1.86 &
  1.24 &
  92.05 &
  81.13 &
  98.10 &
  1.45 &
  0.55 &
  84.73 \\
\rowcolor[HTML]{E0E0E0}
\textbf{EAGLE(Ours)} &
  \textbf{89.02} &
  $\textbf{96.14}$ &
  \textbf{1.84} &
  1.21 &
  \textbf{92.42} &
  \textbf{84.77} &
  \textbf{98.54} &
  \textbf{1.18} &
  \textbf{0.42} &
  \textbf{85.68} \\ \bottomrule
\end{tabular}%
\end{threeparttable}}
\caption{\textbf{Comparison results on Ego4D-VQ3D benchmark}. * indicates official improved baseline.}
\label{tab:vq3d}
\end{table}

We propose a novel multi-view aggregation mechanism that fuses the semantic confidence from the VQL-2D branch with the geometric uncertainty derived from the VGGT, enabling robust estimation of the spatial location of the object. The strategy relies on the key assumption that the 3D location of the target remains relatively stable within a short observation window, an assumption we deem reasonable for the majority of human-object interaction scenarios within Ego4D. Specifically, the VQL-2D network returns a visual track $\mathcal{R}$ in the form of bounding boxes during the retrieval stage. The center coordinates $(u_i, v_i)$ of the corresponding segmentation mask $\mathcal{S}_i$ are used as the 2D localization result for the target in that frame. To associate location information across multiple views, we lift these 2D coordinates into 3D space using the aforementioned aligned geometric information. Based on the principle of inverse projection from the standard pinhole camera model, the corresponding 3D coordinates $[\mathcal{X}_i, \mathcal{Y}_i, \mathcal{Z}_i]^T$ are computed as:$[\mathcal{X}_i, \mathcal{Y}_i, \mathcal{Z}_i, 1]^T = (\mathbf{T}_\eta \mathcal{T}_i) \mathcal{D}_i(u_i, v_i) \mathcal{K}_i^{-1} [u_i, v_i, 1]^T,$
\begin{figure}[tbp]
\centering
\includegraphics[width=0.47\textwidth]{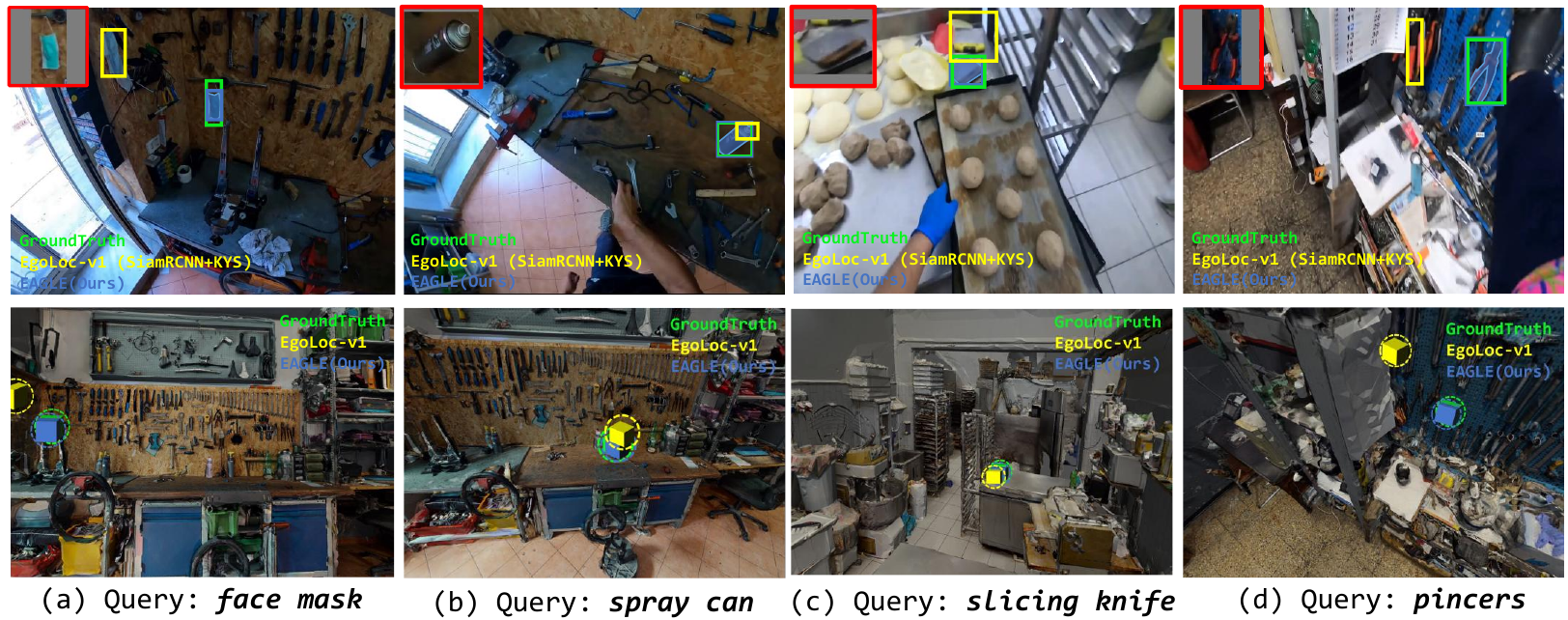}
\caption{\textbf{Visualization of 2D responses and 3D localization}. We back-projected the 2D response predictions and query locations into 3D space. Groundtruth, EAGLE, and EgoLoc-V1, were compared. We don't know the size and rotation of the 3D bbox during the prediction. However, for visualization purposes, we utilize the size and rotation from the ground truth annotations and treat the predicted 3D location as the center of the 3D bbox.}
\label{fig:vq2d_3d}
\end{figure}
where $\mathcal{T}_i$ and $\mathcal{K}_i$ correspond to the camera extrinsics and intrinsics, respectively. Given that single-view localization results are susceptible to various factors, we propose a multi-view aggregation process to obtain an accurate and robust 3D location of the target. The core idea is to perform a weighted average of the 3D coordinates contributed by each view, where the weights are jointly determined by the quality of the segmentation results from the 2D branch and the reliability of the 3D reconstruction. To this end, we define a fused weight $\mathcal{FW}_i=s_{conf}^i \cdot g_{conf}^i$, which is obtained by multiplying two orthogonal confidence components: the semantic confidence $s_{conf}^i$ and the geometric confidence $g_{conf}^i$. This weight attains a higher value only when both the 2D segmentation quality and the 3D geometric reconstruction are reliable, thus ensuring the robustness of the aggregation process. $s_{conf}^i$ is used to evaluate the quality of the 2D segmentation results from VQL-2D. Since the VQL-2D retrieval results are directly reflected by the segmentation results, the semantic confidence comprehensively measures the localization clarity by considering the pixel probabilities within the segmentation mask $\mathcal{S}$. Specifically, we define three sub-metrics: the average probability $\mathbf{P}_{av}$, the maximum probability $\mathbf{P}_{max}$, and the average probability of pixels above a specific threshold $\lambda$, denoted as $\mathbf{P}_{\lambda}$. These are defined as follows:
\begin{equation}
\begin{cases}
\mathbf{P}_{av}= \frac{1}{|\mathcal{S}_i|}\sum_{(u,v)\in \mathcal{S}_i}pr_i^{(u,v)}, \\
\mathbf{P}_{\lambda}= \frac{1}{n}\sum_{\substack{(u,v)\in \mathcal{S}_i, \ pr_i^{(u,v)}>\lambda}} pr_i^{(u,v)}, \\
\mathbf{P}_{max}= \max_{(u,v)\in \mathcal{S}_i}pr_i^{(u,v)},
\end{cases}
\end{equation}
where $pr_i^{(u,v)}$ is the predicted probability that pixel $(u,v)$ belongs to the target, $|\mathcal{S}_i|$ is the total number of pixels in the mask, and $n$ is the number of pixels within the mask whose probability exceeds the threshold $\lambda$. The final semantic confidence is a linear combination of these three metrics: $s_{conf}^i = \varphi \mathbf{P}_{av} + \psi \mathbf{P}_\lambda + \mu \mathbf{P}_{max}$, where $\varphi$, $\psi$, and $\mu$ are hyperparameters, all set to 1/3 in our experiments. The geometric confidence $g_{conf}(i)$ is directly derived from the predicted uncertainty of VGGT. We extract the depth uncertainty value $\tau_i = \mathcal{U}_i(u_i, v_i)$ at the target's center point $(u_i, v_i)$ and convert it into a normalized confidence score $g_{conf}^i= \exp(-\zeta \tau _i)$,
where $\zeta$ is a hyperparameter used to adjust the influence of the uncertainty. This confidence ensures that the contribution of a point to the final aggregation is effectively suppressed when VGGT lacks confidence in its depth prediction for that point. We apply the fused weights $\mathcal{FW}_i$ to multi-view aggregation to obtain a final, aggregated 3D world coordinate for the target, denoted as $[\hat{\mathcal{X}},\hat{\mathcal{Y}},\hat{\mathcal{Z}}]^T$:
\begin{equation}
[\hat{\mathcal{X}}, \hat{\mathcal{Y}}, \hat{\mathcal{Z}}]^T = \frac{\sum_{i=1}^N \mathcal{FW}_i \cdot [\mathcal{X}_i, \mathcal{Y}_i, \mathcal{Z}_i]^T}{\sum_{i=1}^N \mathcal{FW}_i}.
\end{equation}
Finally, we transform this aggregated result back into the camera coordinate system of each specific frame, thus obtaining a 3D relative displacement vector $\delta_i$ with respect to the current observing camera: $\delta_i = (\mathbf{T}_\eta \mathbf{T}_{i})^{-1}[\hat{\mathcal{X}}, \hat{\mathcal{Y}}, \hat{\mathcal{Z}}, 1]^{T}=(\Delta\mathcal{X},\Delta\mathcal{Y},\Delta\mathcal{Z})$. $\delta_i$ represents the final 3D localization result output by our framework for the $i$-th frame, and it implicitly incorporates the consistency constraints derived from multi-view information.

\section{Experiments}
\subsection{Implementation}
Video clips undergo preprocessing, which includes uniform scaling to a $448 \times 448$ resolution via cropping along the longer side and subsequent zero-padding. A Laplacian operator with a window size of 100 is then applied to filter frames. Both training and inference phases involve partitioning the videos into fixed-length clips. The backbone comprises a pre-trained ViT\cite{dinov2}, fine-tuned on EgoTracks and subsequently frozen. The segmentation branch is trained using the Ego4D, EgoTracks\cite{egotracks}, and VISOR\cite{visor} datasets. For the Ego4D and EgoTracks, which provide only bounding box annotations, we leverage SAM to generate segmentation masks, treating them as ground truth. Within the AMM, the meta-learner employs $N_{init}^{train}=10$ and $N_{update}^{train}=3$ for the initial phase and historical frames, respectively. The memory bank size ($\mathcal{O}_{MAX}$) for $\mathcal{O}_{AMM}$ and $\mathcal{O}_{GLM}$ is set to 50. The AdamW is employed for 25,000 iterations, with a peak learning rate of 0.0025 and a weight decay of 0.05, using a linear learning rate scheduler with 2,500 warm-up iterations. We utilize the 1B version of VGGT. We conduct experiments on GTX4090 GPUs.

\noindent\textbf{Training}. We construct a training sequence $\mathcal{V}_{tr}$ by randomly sampling $N$ frames. The first frame initializes the memory banks $\mathcal{O}_{AMM}$ and $\mathcal{O}_{GLM}$. Subsequently, we only update the segmentation branch parameters $\sigma$ while keeping the tracking parameters frozen. The model is optimized with a total loss $\mathcal{L}_{total}=\mathcal{L}_{seg}+\rho\,\mathcal{L}_{tck}$, where $\mathcal{L}_{seg}$ is the Lovász loss, defined as $\sum_{j=1}^{J-1}\mathcal{L}_\sigma\bigl(\mathbf{D}(\mathcal{A}_\sigma^{j-1}(\mathcal{F}_j)+\kappa_\theta(c(\mathcal{F}_j))),\mathcal{M}_j\bigr)$, and $\mathcal{L}_{tck}$ is a hinge loss, formulated as $\sum_{j=1}^{J-1}\tfrac{1}{N_{iter}}\sum_{i=0}^{N_{iter}}\mathcal{L}_c(\mathcal{F}_i*c,G_i)$. $\rho$ is weighting factor.

\noindent\textbf{Inference}. The video is first processed in clips, and the predictions are concatenated. The dual memory banks are initialized with the visual query, updated continuously for the first 100 historical frames, and then every 25 frames thereafter. The initial query sample is augmented using flipping, translation, and blurring. We use the mask confidence score, $s_{conf}$, as the temporal score. After applying a 5-frame median filter, we set a threshold at $0.8 \times \max(s_{conf})$. The last interval exceeding this threshold is output as the final temporal localization for VQL-2D.

\subsection{Comparison to the state-of-the-art}
We evaluate EAGLE on Ego4D-VQ, the unique publicly available benchmark for VQL. As shown in Table.\ref{tab:vq2d}, On the test set, our method surpasses the previous best, RELOCATE, by 6.9\% in tAP$_{25}$, 14.3\% in stAP$_{25}$, 5.8\% in Rec, and 4.3\% in Succ. On the validation set, we achieve performance gains of 14.6\%, 27.3\%, 3.1\%, and 5.6\% in the same metrics, respectively, while maintaining a comparable inference speed. Qualitative comparisons with VQLoc, presented in Figure.\ref{fig:quality}, highlight EAGLE's superior performance, robustness, and generalization capabilities in challenging egocentric scenarios. For the VQL-3D, EAGLE's performance is detailed in Table.\ref{tab:vq3d}. Compared to EgoLoc-V1, EAGLE improves Succ and QwP by 0.4\% on the test set, while reducing L2 and Angle errors by 1.1\% and 2.4\%. The improvements on the validation set are more substantial, with gains of 4.5\% in Succ, 0.4\% in Succ*, and 1.12\% in QwP, alongside significant reductions of 18.6\% in L2 and 23.6\% in Angle errors. These advancements are attributed to VGGT's comprehensive estimation of camera pose and depth, which allows for the utilization of a greater number of camera poses in the computation. Qualitative results are visualized in Figure\ref{fig:vq2d_3d}.
\subsection{Ablation Analysis}
\textbf{Impact of AMM}. To investigate the impact of each component within AMM, we conduct an ablation study with five variants, as detailed in Table.\ref{tab:ablationamm}: 
(i) $STA\rightarrow SAM$: We replace SAM with a pre-trained STA\cite{sta} model during the pseudo-mask generation phase. (ii) w/o $\mathcal{P}_{\theta}$: We ablate the pseudo label modulator network, utilizing only binary mask information. (iii) w/o $\mathcal{W}_{\theta}$: We remove the target re-weighting network. 
(iv) $\mathcal{Q}_0!\rightarrow \phi$: Within $\mathcal{O}_{AMM}$, all queues except for the initial query's are set to a FIFO scheme, preventing the initial query from being replaced. (v) w/o $\mathcal{O}_{AMM}$: We completely remove the $\mathcal{O}_{AMM}$ module. The results indicate that the removal of any component adversely affects performance. Most notably, ablating the $\mathcal{O}_{AMM}$ leads to the most significant performance degradation, with drops of 11.9\% in tAP$_{25}$, 40\% in stAP$_{25}$, 23.4\% in Rec, and 12.2\% in Succ. This underscores the critical role of the memory bank in stabilizing target retrieval. The second most impactful change is the substitution of SAM with STA, which demonstrates that SAM generates higher-quality pseudo-labels containing more discriminative information for retrieval, thereby enabling more precise segmentation.
\begin{table}[t]
\centering
\scriptsize 
\setlength{\tabcolsep}{2.5pt} 
\begin{tabular}{@{}ccccc|cccc@{}}
\toprule
\multirow{2}{*}{\shortstack{STA\\$\rightarrow$SAM}} &
  \multirow{2}{*}{\shortstack{w/o\\$\mathcal{P}_\theta$}} &
  \multirow{2}{*}{\shortstack{w/o\\$\mathcal{W}_\theta$}} &
  \multirow{2}{*}{\shortstack{$Q_0$\\!$\rightarrow\phi$}} &
  \multirow{2}{*}{\shortstack{w/o\\$\mathcal{O}_{AMM}$}} &
  \multicolumn{4}{c}{Ego4D-VQ2D Validation Set} \\ \cmidrule(l){6-9} 
  & & & & & tAP$_{25}\uparrow$ & stAP$_{25}\uparrow$ & Rec.(\%)$\uparrow$ & Succ.(\%)$\uparrow$ \\ \midrule $\checkmark$ & - & - & - & - & 0.44 & 0.30 & 45.19 & 57.28 \\
- & $\checkmark$ & - & - & - & 0.46 & 0.33 & 51.08 & 60.22 \\
- & - & $\checkmark$ & - & - & 0.47 & 0.32 & 51.55 & 61.01 \\
- & - & - & $\checkmark$ & - & 0.46 & 0.32 & 50.61 & 60.57 \\
- & - & - & - & $\checkmark$ & 0.42 & 0.30 & 42.22 & 54.62 \\
\rowcolor{gray!20}
- & - & - & - & - & \textbf{0.47} & \textbf{0.42} & \textbf{52.09} & \textbf{61.29} \\ \bottomrule
\end{tabular}
\caption{\textbf{Ablation study of AMM on Ego4D-VQ2D validation set.} The final model configuration is highlighted in gray, a convention adopted hereinafter.}
\label{tab:ablationamm}
\end{table}

\noindent\textbf{Impact of GLM}. We also designed two variants to investigate the impact of the GLM: (i) $\mathcal{Q}_0\to\phi$: In $\mathcal{O}_{GLM}$, all queues are updated using a FIFO scheme, allowing the initial query to be replaced; and (ii) w/o $\mathcal{O}_{GLM}$: The $\mathcal{O}_{GLM}$ is completely removed. The experimental results show that removing $\mathcal{O}_{GLM}$ has the most pronounced impact on performance, leading to decreases of 10.6\% in tAP$_{25}$, 29.8\% in stAP$_{25}$, 18.7\% in Rec, and 13.4\% in Succ. This demonstrates that the historical visual cues provided by $\mathcal{O}_{GLM}$ is crucial for stable, long-term tracking within the discriminative branch. Interestingly, we find that the optimal update strategy for the GLM's memory is the inverse of that for the AMM; updating GLM's initial query along with the other memory sequences degrades performance. We attribute this to the distinct feature granularities handled by the two branches. The segmentation branch processes pixel-level information and is highly sensitive to contour and scale consistency; any accumulated error directly degrades the mask, so memory must be updated synchronously to reflect the latest appearance. In contrast, the GLM-guided discriminative tracking branch operates on region-level geometric descriptors and focuses on the target’s identity (“what it is”), which gives it greater tolerance to minor appearance drift. Retaining the initial query as a fixed identity anchor effectively mitigates drift caused by prolonged occlusion or appearance changes, thereby maintaining tracking stability.
\begin{table}[]
\centering
\footnotesize
\begin{tabular}{@{}cc|cccc@{}}
\toprule
& & \multicolumn{4}{c}{Ego4D-VQ2D Validation Set} \\ \cmidrule(l){3-6} 
\multirow{-2}{*}{\shortstack{$\mathcal{Q}_0$ \\ $\rightarrow \phi$}} & 
\multirow{-2}{*}{\shortstack{w/o \\ $\mathcal{O}_{GLM}$}} & 
tAP$_{25}$↑ & stAP$_{25}$↑ & Rec.(\%)↑ & Succ.(\%)↑ \\ 
\midrule
$\checkmark$ & - & 0.44 & 0.29  & 46.35  & 57.73 \\
- & $\checkmark$ & 0.42 & 0.28 & 42.36  & 53.09  \\
\rowcolor{gray!20}
\textbf{-}             & \textbf{-}                  & \textbf{0.47} & \textbf{0.42} & \textbf{52.09} & \textbf{61.29} \\ \bottomrule
\end{tabular}
\caption{\textbf{Ablation study of GLM on Ego4D-VQ2D validation set}}
\label{tab:ablationglm}
\end{table}
\begin{table}[t!]
\centering
\scriptsize
\begin{tabular}{@{}c|cccc@{}}
\toprule
                           & \multicolumn{4}{c}{Ego4D-VQ2D Validation Set} \\ \cmidrule(l){2-5} 
\multirow{-2}{*}{Backbone@Resolution} & tAP$_{25}\uparrow$      & stAP$_{25}\uparrow$     & Rec.(\%)$\uparrow$       & Succ.(\%)$\uparrow$      \\ \midrule
DINOv2-ViT-S/14@224        & 0.38       & 0.26       & 46.69     & 50.29     \\
DINOv2-ViT-B/14@224        & 0.39       & 0.27       & 47.25     & 51.21     \\
\rowcolor{gray!20}
DINOv2-ViT-B/14@448        & \textbf{0.47}        & \textbf{0.42}       & \textbf{52.09}     & \textbf{61.29}     \\
CLIP-ViT-B/16@448          & 0.36       & 0.29       & 49.96     & 52.45     \\ \bottomrule
\end{tabular}
\caption{\textbf{Ablation study of backbone and input resolution}}
\label{tab:ablationbb}
\end{table}
\begin{table}[t]
\centering
\resizebox{\columnwidth}{!}{%
\tiny
\begin{tabular}{@{}ccccc|ccccc@{}}
\toprule
& \multicolumn{3}{c}{$s_{conf}$} & & \multicolumn{5}{c}{Ego4D-VQ3D Validation   Set} \\ \cmidrule(lr){2-4} \cmidrule(l){6-10} 
\multirow{-2}{*}{\shortstack{Last-Resp.\\(Baseline)}} & $\mathbf{P}_{av}$ & $\mathbf{P}_\lambda$ & $\mathbf{P}_{max}$& \multirow{-2}{*}{$g_{conf}$} & Succ.(\%)$\uparrow$ & Succ.*(\%)$\uparrow$& L2$\downarrow$ & Angle$\downarrow$& QwP\%$\uparrow$ \\ \midrule
$\checkmark$ & -   & -  & -  & -          & 79.26    & 92.35   & 1.54   & 0.65    & 85.68   \\
-   & $\checkmark$ & -     & -       & - & 79.95  & 92.76   & 1.52   & 0.63    & 85.68   \\
-  & -& $\checkmark$     & -   & -       & 80.01  & 93.25   & 1.48   & 0.57& 85.68   \\
- & -  & - & $\checkmark$ & -    & 81.65    & 94.53   & 1.34   & 0.51    & 85.68   \\
- & $\checkmark$& $\checkmark$     & -  & -  & 81.92    & 96.75   & 1.28   & 0.49    & 85.68   \\
-  & $\checkmark$  & -     & $\checkmark$ & - & 82.22    & 96.49   & 1.25   & 0.45    & 85.68   \\
-  & - & $\checkmark$     & $\checkmark$  & -  & 82.29    & 96.88   & 1.26   & 0.45    & 85.68   \\
-  & -       & -     & -       & $\checkmark$ & 80.33    & 94.65   & 1.51   & 0.55    & 85.68   \\
\rowcolor{gray!20}
-     & $\checkmark$  & $\checkmark$     & $\checkmark$       & $\checkmark$ & \textbf{84.77}    & \textbf{98.54}   & \textbf{1.18}   & \textbf{0.42}    & \textbf{85.68}   \\ \bottomrule
\end{tabular}%
}
\caption{\textbf{Ablation study of multi-view aggregation function on the Ego4D-VQ3D validation set.}}
\label{tab:ablationaggregation}
\end{table}

\noindent\textbf{Impact of backbone size and input resolution}. We conducted four variants:(i) DINOv2-ViT-S/14 with an input resolution of $448 \times 448$; (ii) DINOv2-ViT-B/14 with $224 \times 224$; (iii) DINOv2-ViT-B/14 with $448 \times 448$; and (iv) CLIP-ViT-B/16 \footnote{https://github.com/openai/CLIP} with $448 \times 448$. The results indicate that both backbone capacity and input resolution correlate positively with performance, whereas lowering the resolution leads to a substantial decline in accuracy. We therefore adopt the best-performing model, DINOv2-ViT-B/14@448.

\noindent\textbf{Impact of multi-view aggregation function}. Table.\ref{tab:ablationaggregation} presents the ablation study on variants of semantic and geometric confidence. The results indicate that each semantic confidence metric ($\mathbf{P}_{av}$, $\mathbf{P}_\lambda$, and $\mathbf{P}_{max}$), when used individually, outperforms the baseline(only use last response). This superiority stems from their distinct approaches to handling mask pixel confidence: $\mathbf{P}_{av}$ mitigates the impact of missed detections by comprehensively evaluating the quality of region coverage; $\mathbf{P}_\lambda$ reduces background interference by filtering out low-confidence pixels via a threshold; and $\mathbf{P}_{max}$ focuses on high-confidence pixels to decrease sensitivity to noise and false positives, thereby significantly improving the L2 and Angle metrics. Notably, these three methods are complementary when combined. Furthermore, geometric confidence reflects the reliability of pixels in 3D space. Its integration with semantic confidence creates a synergistic effect, where their complementary strengths jointly enhance the precision of spatial localization.
\section{Conclusions}
In this paper, we addressed critical limitations in VQL by moving beyond “detect-then-track” methods. We introduced a novel framework inspired by biological memory consolidation, which synergizes segmentation and tracking to build a robust, long-term episodic memory. We establish the online memory bank that actively filters and stores high-confidence samples, enabling stable retrieval through significant appearance variations and environmental distractors. Furthermore, we unified 2D and 3D localization into an efficient pipeline. Our work achieves SOTA performance on the Ego4D-VQ benchmark, demonstrating its potential as a new baseline.
\section{Acknowledgments}
This work was supported by the National Natural Science Foundation of China(Grant:61672128) and from the Dalian Key Field Innovation Team Support Plan(Grant:2020RT07).
\bibliography{aaai2026}
\clearpage
\onecolumn
\appendix
\section{Supplementary Material}
\noindent\textbf{Code are available at \url{https://github.com/cyfedu-dlut/EAGLE_VQL}}
\subsection{Preliminaries}
\label{appendix:preliminaries}
\textbf{EAGLE in VQL task}. Given a visual query $\mathcal{Q}$, specified as an image crop of target, and a set of video frames $\mathcal{V}=\{I_i\}_{i=0}^{N-1}$ consisting of $N$ frames, our objective is to localize and track the object throughout the video, and further recover its 3D position in the world coordinate. Firstly, for each video frame $I_i$, we employ the off-the-shelf model (e.g., VGGT) to predict both the camera pose $\mathcal{T}_i$ and depth map $d_i$. Next, we input $\mathcal{Q}$ and $\mathcal{V}$ into a VQL-2D model, which retrieves the target and generates a sequence of 2D bounding boxes $rt=\{rt_s,rt_{s+1},\cdots,rt_e\}$, where $s$ and $e$ denote the start and end frame indices, respectively, and each $rt_i$ represents the bounding box containing the query in frame $I_i$. Subsequently, for each response frame $I_i$, we use $\mathcal{T}_{I_i}$ and $d_{I_i}$ to back-project the center of the 2D location to the 3D scene, obtaining a set of predicted 3D coordinates $[x_{I_i},y_{I_i},z_{I_i}]$. By aggregating these coordinates across all response frames, we recover the object’s 3D position $[\hat{x},\hat{y},\hat{z}]$ in the world coordinate. Finally, the relative displacement $\delta$ is obtained by projecting $\mathcal{T}_{I_i}$ relative to the query frame.
\subsection{Derivation of meta-learner iteration steps in AMM}
\label{appendix:segderivations}
In this section, we derive the steepest descent update rule (Eq.(\ref{equ:steepupdate})) employed in our meta-learner to minimize the loss function (Eq.(\ref{equ:lossseg})). For clarity, we first reformulate the loss in matrix-vector notation, which allows us to derive closed-form expressions for both the vectorized gradient $\hat{g}$ and the optimal step size $\alpha$. For a feature map $\mathcal{F}_i\in\mathbb{R}^{H\times W\times C}$ and a convolution kernel $\sigma\in\mathbb{R}^{K\times K \times C\times D}$, we express their convolution using matrix multiplication as $mp(\mathcal{F}_i*\sigma)=\mathbf{F}_i\hat{\sigma}$. Here, $mp$ denotes the flattening operation, $\hat{\sigma}=mp(\sigma) \in\mathbb{R}^{K^2CD}$, and $\mathbf{F}_i \in\mathbb{R}^{HWD\times K^2CD}$ is the matrix representation of $[\mathcal{F}_i*]$. Furthermore, we define $e_i=mp(\mathcal{P}_\theta(\mathcal{M}_i))\in \mathbb{R}^{HWD}$ as the vectorized form of the label encoding. The matrix $W_i=diag(mp(\mathcal{W}_\theta(\mathcal{M}_i)))\in \mathbb{R}^{HWD\times HWD}$ is a diagonal matrix corresponding to the target re-weighting network. Now, we can write Eq.(\ref{equ:lossseg}) in matrix form as:
\begin{equation}\mathcal{L}(\hat{\sigma})=\frac{1}{2}\sum_{i}\left\|W_{i}(\mathbf{F}_i\hat{\sigma}-p_{i})\right\|^{2}+\frac{\delta}{2}\left\|\hat{\sigma} \right\|^{2}\end{equation}
Consequently, we update the parameters in the gradient direction $\hat{g}^i$ with a step size $\alpha^i$, such that $\hat{\sigma}_{i+1}=\hat{\sigma}_i-\alpha^i\hat{g}^i$. By setting $\hat{rd}(\hat{\sigma})=W_i(\mathbf{F}_i\hat{\sigma}-e_i)$, we can apply the chain rule as follows:
\begin{equation}
\label{equ:chainder}
\hat{g}=\nabla \mathcal{L}(\hat{\sigma}_d)=\sum_i\left(\frac{\partial\hat{rd}_i}{\partial\hat{\sigma}}\right)^\mathrm{T}\hat{rd}_i(\hat{\sigma})+\delta\hat{\sigma}=\sum_i\mathbf{F}_i^\mathrm{T}W_i^2\left(\mathbf{F}_i\hat{\sigma}-e_i\right)+\delta\hat{\sigma}\end{equation}.
The gradient is now computed as follows:
\begin{equation}\begin{aligned}\hat{g}&=\sum_i\mathbf{F}_i^\mathrm{T}W_t^2\left(\operatorname{mp}(\mathcal{F}_i*\sigma)-e_i\right)+\delta\operatorname{mp}(\sigma)\\&=\operatorname{mp}\left(\sum_i\mathcal{F}_i*^\mathrm{T}\mathcal{W}_\theta(\mathcal{M}_i)^2\cdot(\mathcal{F}_i*\sigma-\mathcal{P}_\theta(\mathcal{M}_i))+\delta\sigma\right)\end{aligned}\end{equation}
where the transposed convolution $\mathcal{F}_*^\mathrm{T}$ corresponds to the matrix multiplication by $\mathbf{F}_i^\mathrm{T}$. Thus,
\begin{equation}g=\sum_i\mathcal{F}_i*^\mathrm{T}\left(\mathcal{W}_\theta(\mathcal{M}_i)^2\cdot\left(\mathcal{F}_i*\sigma-\mathcal{P}_\theta(\mathcal{M}_i)\right)\right)+\delta\sigma\end{equation}
We compute the step size $\alpha^i$ that minimizes $\mathcal{L}$ in the current gradient direction $g^i$:
\begin{equation}\alpha^i=\arg\min_\alpha \mathcal{L}(\sigma_i-\alpha\hat{g}^i)\end{equation}
Since the loss function is convex, it possesses a unique global minimum, which is obtained by solving for the stationary point:
\begin{equation}\frac{\mathrm{d}\mathcal{L}(\hat{\sigma}_{i}-\alpha\hat{g}^{i})}{\mathrm{d}\alpha}=0\end{equation}
By setting $\vartheta =\hat{\sigma}_i-\alpha\hat{g}^i$ and applying the chain rule in conjunction with Eq.(\ref{equ:chainder}), we obtain:
\begin{equation}
\begin{aligned}
0=\frac{\mathrm{d}\mathcal{L}(\vartheta)}{\mathrm{d}\alpha}&\begin{aligned}=\left(\frac{\mathrm{d} \vartheta}{\mathrm{d}\alpha}\right)^\mathrm{T}\nabla_\vartheta \mathcal{L}(\vartheta)\end{aligned}\\&=(\hat{g}^i)^\mathrm{T}\left(\sum_i\mathbf{F}_i^\mathrm{T}W_i^2\left(\mathbf{F}_i(\hat{\sigma}^i-\alpha\hat{g}^i)-e_i\right)+\delta(\hat{\sigma}_i-\alpha\hat{g}^i)\right)\\&=(\hat{g}^i)^\mathrm{T}\hat{g}^i-\alpha(\hat{g}^i)^\mathrm{T}\left(\sum_i\mathbf{F}_i^\mathrm{T}W_i^2\mathbf{F}_i\hat{g}^i+\delta\hat{g}^i\right)\\&=\|\hat{g}^i\|^2-\alpha\left(\sum_i\|W_i\mathbf{F}_i\hat{g}^i\|^2+\delta\|\hat{g}^i\|^2\right).
\end{aligned}
\end{equation}
Therefore, the step size is computed as follows:
\begin{equation}\alpha=\frac{\left\|\hat{g}^i\right\|^2}{\sum_i\left\|W_i\mathbf{F}_i\hat{g}^i\right\|^2+\delta\left\|\hat{g}^i\right\|^2}\end{equation}
where $\left\|\hat{g}^i\right\|^2=\left\|g^i\right\|^2$ and $\left\|W_i\mathbf{F}_i\hat{g}^i\right\|^2=\left\|mp(\mathcal{W}_\theta(\mathcal{M}_i)\cdot \mathcal{F}_i*g^i)\right\|^2$. The step size calculation thus becomes:
\begin{equation}\alpha=\frac{\left\|g^i\right\|^2}{\sum_i\left\|\mathcal{W}_\theta(\mathcal{M}_i)\cdot(\mathcal{F}_i*g^i)\right\|^2+\delta\left\|g^i\right\|^2}\end{equation}
\subsection{Derivations of closed-form expression in GLM}
\label{appendix:trackderivations}
In this section, we derive the closed-form solution for the loss function presented in Eq.(\ref{equ:1}) of the main paper.
\begin{equation}
\label{equ:s1}
   \mathcal{L}(c) = \frac{1}{|\mathcal{O}_{GLM}|} \sum_{(\mathcal{F}(\cdot),G(\cdot))\in\mathcal{O}_{GLM}} \|\mathcal{H}(\mathcal{H}_\mathcal{J}, G)\|^2 + \|\lambda c\|^2. 
\end{equation}
Here, $\mathcal{H}(\mathcal{H}_\mathcal{J}, G)$ denotes the spatial residual between the predicted score map $\mathcal{H}_\mathcal{J} = \mathcal{F} * c$ and its corresponding Gaussian label $G$. The training set is given by $\mathcal{O}_{GLM} = \{(\mathcal{F}_i, G_i)\}_{i=1}^n$. The residual function $\mathcal{H}(\mathcal{H}_\mathcal{J}, G)$ is defined as (reiterated from Eq.(\ref{equ:2}) in the main paper):
\begin{equation}
\label{equ:s2}
    \mathcal{H}(\mathcal{H}_\mathcal{J}, G) = sw_G \cdot (\mathcal{S}_i \mathcal{H}_\mathcal{J} + (1-\mathcal{S}_i) \max(0, \mathcal{H}_\mathcal{J}) - G_i)
\end{equation}
The gradient of the loss function (Eq.(\ref{equ:s1})) with respect to the filter coefficients $c$, denoted as $\nabla \mathcal{L}(c)$, is computed as follows:
\begin{equation}
\label{equ:s3}
\nabla L(c)=\frac{2}{|\mathcal{O}_{GLM}|}\sum_{(\mathcal{F}(\cdot),G(\cdot))\in \mathcal{O}_{GLM}}\left(\frac{\partial \mathcal{H}(\mathcal{H}_\mathcal{J},G)}{\partial c}\right)^{\mathrm{T}}\mathcal{H}(\mathcal{H}_\mathcal{J}, G) +2\lambda^{2}c.
\end{equation}
Here, $\frac{\partial \mathcal{H}(\mathcal{H}_\mathcal{J},G)}{\partial c}$ represents the Jacobian matrix of the residual function (Eq.(\ref{equ:s2})) with respect to the filter coefficients $c$. Leveraging Eq.(\ref{equ:s2}), we obtain:
\begin{equation}
\label{equ:s4}
\frac{\partial \mathcal{H}(\mathcal{H}_\mathcal{J},G)}{\partial f}=\mathrm{diag}(sw_G\mathcal{S})\frac{\partial \mathcal{H}_\mathcal{J}}{\partial c}+\mathrm{diag}\left((1-\mathcal{S})\cdot\mathbb{1}_{ \mathcal{H}_\mathcal{J}>0}\right)\frac{\partial \mathcal{H}_\mathcal{J}}{\partial c}=\mathrm{diag}(Q_{G})\frac{\partial \mathcal{H}_\mathcal{J}}{\partial c}.\end{equation}
where $\mathrm{diag}(Q_{G})$ denotes a diagonal matrix containing all the elements of $Q_{G}$. Furthermore, $Q_{G}$ is calculated using only element-wise operations as $Q_{G} = sw_G \mathcal{S} + (1-\mathcal{S}) \mathbb{1}_{\mathcal{H}_\mathcal{J}>0}$, where $\mathbb{1}_{\mathcal{H}_\mathcal{J}>0}$ is an indicator function that equals 1 if $\mathcal{H}_\mathcal{J}$ is positive, and 0 otherwise. By substituting Eq.(\ref{equ:s4}) into Eq.(\ref{equ:s3}), we arrive at the final expression:
\begin{equation}
\label{equ:s5}\nabla \mathcal{L}(c)=\frac{2}{|\mathcal{O}_{GLM}|}\sum_{(\mathcal{F}(\cdot),G(\cdot))\in\mathcal{O}_{GLM}}\left(\frac{\partial \mathcal{H}_\mathcal{J}}{\partial c}\right)^{\mathrm{T}}( Q_{G} \cdot \mathcal{H}(\mathcal{H}_\mathcal{J},G))+2\lambda^{2}c.\end{equation}
Here, $\odot$ denotes the element-wise product. The multiplication by the transposed Jacobian, $\left(\frac{\partial \mathcal{H}_\mathcal{J}}{\partial c}\right)^{\mathrm{T}}$, corresponds to the backpropagation of the input $Q_{G} \odot \mathcal{H}(\mathcal{H}_\mathcal{J},G)$ through the convolutional layer $c \mapsto \mathcal{F}*c$. This is implemented as a transposed convolution with respect to $\mathcal{F}$ using standard operations in deep learning libraries such as PyTorch. Consequently, the closed-form expression in Eq. (\ref{equ:s5}) can be readily implemented using standard functionalities.
\subsection{More architecture details}
\label{app:modeldetails}
\subsubsection{Coordinate Alignment.}
\label{appendix:coordinate}
The 3D information predicted by VGGT is defined in an arbitrary yet self-consistent coordinate, typically anchored to the first frame. To enable evaluation on the VQL-3D benchmark, which uses the Matterport Scan coordinate, an alignment between these two frames of reference is imperative. To this end, we first convert the native geometric outputs of VGGT into the standard COLMAP format, which encapsulates the camera poses and a sparse 3D point cloud generated by back-projecting the depth maps. Subsequently, we employ a standard Sim(3) alignment procedure to solve for a 7-DoF transformation matrix, $\mathbf{T}_\eta \in \text{Sim}(3)$ (scale, rotation, and translation), that minimizes the error between the VGGT-predicted point cloud and the corresponding points in the ground-truth coordinate system. This is formulated as a least-squares problem:
\begin{equation}
\mathbf{T}_\eta = \underset{\mathbf{T} \in \text{Sim}(3)}{\arg\min} \sum_{j} || \mathbf{T} \cdot \mathbf{pc}_{j}^{\text{vggt}} - \mathbf{pc}_{j}^{\text{ms}} ||^2
\end{equation}
where $\mathbf{T}$ is a candidate transformation matrix being optimized over the Sim(3) space. The terms $\mathbf{pc}_{j}^{\text{vggt}}$ and $\mathbf{pc}_{j}^{\text{ms}}$ denote the $j$-th pair of 3D corresponding points extracted from the VGGT point cloud and the ground-truth coordinate system, respectively. This alignment process is performed dynamically for each test sequence to ensure optimal fitting. Upon obtaining $\mathbf{T}_\eta$, we apply it to all camera extrinsics predicted by VGGT, thereby unifying the entire reconstruction into the canonical benchmark coordinate.

\subsection{Training and Inference}
\label{appendix:traininginference}
\noindent\textbf{Training}. We construct the training sequence $\mathcal{V}_{tr}=\{(\mathcal{F}_j, \mathcal{M}_j, G_j)\}_{j=0}^{J-1}$ by randomly sampling $N$ frames from annotated video sequences and sorting them in ascending order according to their frame IDs. Starting with the initial frame $I_0 \in \mathcal{V}$, we store it in the memory banks $\mathcal{O}_{AMM}$ and $\mathcal{O}_{GLM}$. Subsequently, we compute the segmentation loss for the next frame $I_1 \in \mathcal{V}$ using the initialization parameters. Based on the predicted segmentation mask, we update the parameters of the segmentation model to $\sigma_1$, while keeping the tracking model parameters fixed, reflecting the higher update frequency typically required for segmentation accuracy. In contrast, the tracking model is trained on a single frame prediction, enabling the tracking branch to generalize to multiple unseen future frames, thereby ensuring robust target localization. Consequently, the segmentation loss for the entire training sequence $\mathcal{V}_{tr}$ can be formulated as:
\begin{equation}
\mathcal{L}_{\mathrm{seg}}(\theta;\mathcal{V}_{tr})=\sum_{j=1}^{J-1}\mathcal{L}_\sigma\left(\mathbf{D}\left(\mathcal{A}_\sigma^{j-1}(\mathcal{F}_j)+\kappa_\theta\left(c(\mathcal{F}_j)\right)\right),\mathcal{M}_j\right)
\end{equation}
where $\mathcal{L}_\sigma$ denotes the Lovasz segmentation loss \cite{loss}. To further enhance the accuracy of target localization, we also incorporate a classification loss, defined as:
\begin{equation}
\mathcal{L}_{tck}(\theta;\mathcal{V}_{tr})=\sum_{j=1}^{J-1}\left(\frac{1}{N_{\mathrm{iter}}}\sum_{i=0}^{N_{\mathrm{iter}}}\mathcal{L}_c\left(\mathcal{F}_i*c,G_i\right)\right),
\end{equation}
where $\mathcal{L}_c$ represents the hinge loss. Finally, the total loss $\mathcal{L}_{total}$ is defined as a weighted combination of the segmentation and classification losses, with a weighting factor of $\rho$:
\begin{equation}
\mathcal{L}_{total}(\theta;\mathcal{V}_{tr})=\mathcal{L}_{seg}(\theta;\mathcal{V}_{tr})+\rho\cdot\mathcal{L}_{tck}(\theta;\mathcal{V}_{tr}).
\end{equation}

\noindent\textbf{Inference}. The video is divided into clips, and the predicted trajectories and temporal scores from these clips are then concatenated into a continuous prediction sequence, thereby integrating the analysis results from multiple segments. The initial query samples are permanently stored in two memory banks. These memory banks are updated within the first 100 video frames, and subsequently updated every 25 frames. Data augmentation techniques, including vertical flipping, random translation, and blurring, are applied to the initial query samples. Given that bounding box generation primarily relies on the segmentation results, we adopt the segmentation mask confidence score, $s_{conf}$, as the corresponding temporal score. To smooth the temporal scores, a median filter with a window size of 5 is employed. Subsequently, peak detection is performed to identify the maximum peak in the matching scores. Finally, a threshold of 0.8 times the maximum peak is selected to filter the score sequence, and the last time interval in the score sequence that exceeds the threshold is retained as the temporal result for VQL-2D.

\subsection{Additional Ablations and Experiments Details}
\label{appendix:additionablations}
\subsubsection{Segmentation confidence threshold $s_{conf}$.}
\label{appendix:segthreshold}
Building on the generative mask for joint segmentation and tracking, we employ a confidence threshold to filter samples added to the memory bank and those ultimately utilized in the aggregation process for the 3D branch. Table.\ref{tab:confidencethreshold} shows that we conduct ablation experiments with varying confidence thresholds (0.2, 0.4, 0.6, 0.8) on the VQ2D and VQ3D validation sets, revealing a performance trend that initially increases with higher thresholds before subsequently declining. Our findings indicate that a high confidence threshold ($s_{conf} = 0.8$) tends to retain samples with clear targets but is susceptible to drift in low-confidence scenarios (e.g., heavy occlusion), resulting in insufficient memory bank capacity. This causes the model to lack sufficient “historical knowledge” to adapt to changes in subsequent frames, leading the AMM- and GLM-guided dual-branch architecture to fail in capturing sufficiently diverse samples. Consequently, the fusion and aggregation processes, which rely on this threshold, suffer from a lack of rich memory support. Conversely, a low confidence threshold ($s_{conf} = 0.2$) incorporates incomplete and unclear samples, degrading the memory bank’s quality and causing performance decline. We ultimately selected a confidence threshold of 0.6 to balance performance and sample quality. It should be noted that if the global $s_{conf}$ consistently falls below 0.4, the update process is halted, and the system reverts to the initial query.
\begin{table}[]
\centering
\caption{Ablation study of segmentation confidence threshold}
\label{tab:confidencethreshold}
\begin{tabular}{@{}cccccccccc@{}}
\toprule
\multirow{2}{*}{Thresold} & \multicolumn{4}{c}{VQ2D validation Set} & \multicolumn{5}{c}{VQ3D validation Set} \\ \cmidrule(l){2-10} 
    & tAP$_{25}\uparrow$  & stAP$_{25}\uparrow$ & Rec.(\%)$\uparrow$ & Succ.(\%)$\uparrow$ & Succ.(\%)$\uparrow$  & Succ*(\%)$\uparrow$ & L2$\downarrow$   & Angle$\downarrow$ & QwP(\%)$\uparrow$   \\ \midrule
0.2 & 0.21 & 0.18 & 32.22  & 35.11   & 40.26 & 47.11 & 4.42 & 1.72  & 85.68 \\
0.4 & 0.37 & 0.30 & 45.32  & 55.73   & 78.23 & 85.59 & 1.45 & 0.91  & 85.68 \\
0.6 & 0.47  & 0.42 & 52.09  & 61.29   & 84.77 & 98.54 & 1.18 & 0.42  & 85.68 \\
0.8 & 0.45 & 0.41 & 51.77  & 61.03   & 82.69 & 94.56 & 1.21 & 0.45  & 85.68 \\ \bottomrule
\end{tabular}
\end{table}

\subsubsection{SAM settings.}
\label{appendix:sam}
\begin{figure}[!h]
\centering
\begin{minipage}{0.48\textwidth}
    \centering
    \includegraphics[width=\textwidth,height=0.7\columnwidth]{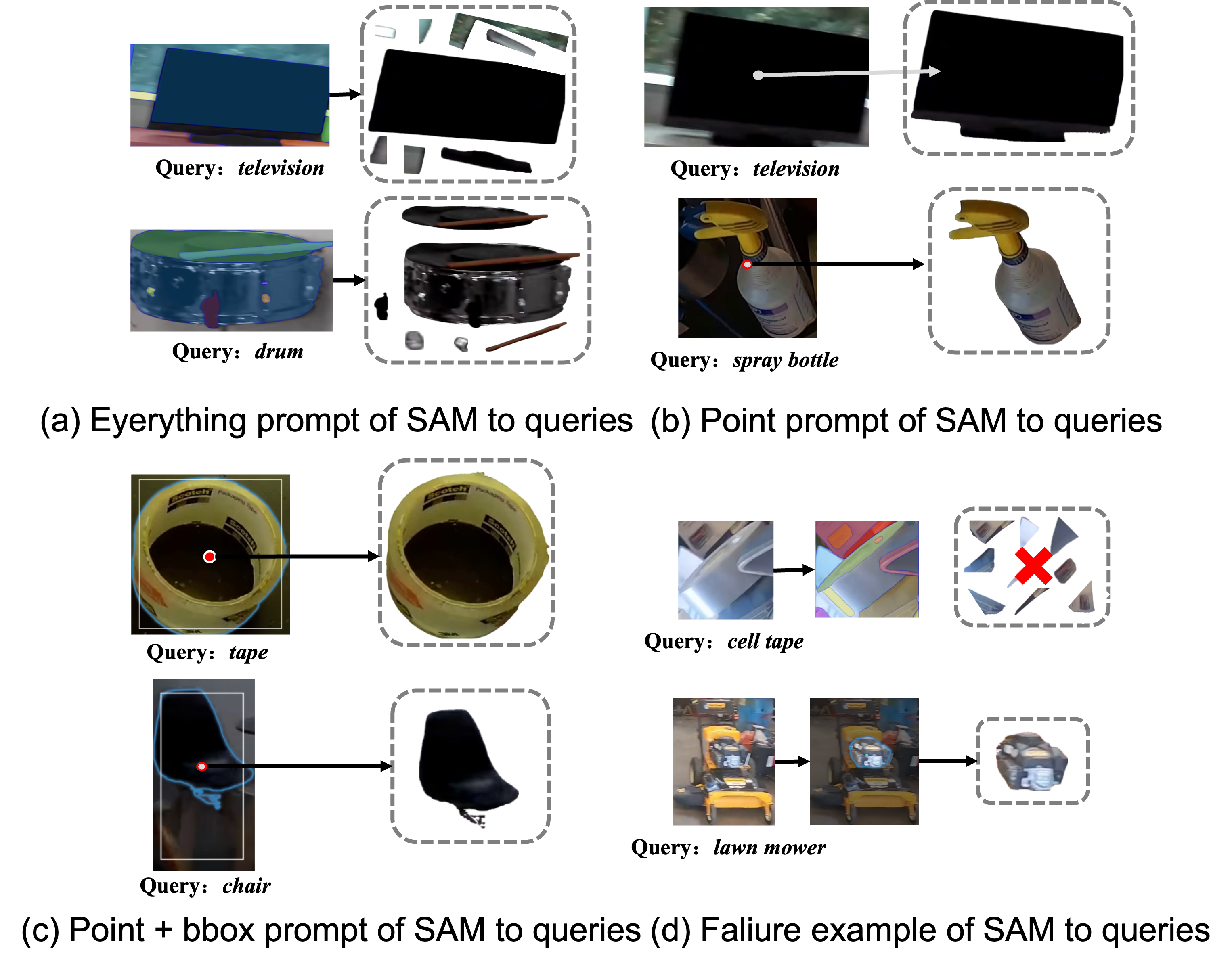}
    \caption{\textbf{Qualitative analysis on SAM pathways}. (a) illustrates segmentation of the largest object based on target pixel proportion using the \textit{Everything} prompt; (b) demonstrates segmentation centered on the image's center point, based on a center-point prior; (c) shows improved segmentation results by combining a \textit{point} prompt with a bounding box sized to 2/3 of the query image; (d) reveals that \textit{Everything} and \textit{point} prompt can sometimes fail, while the method in (c) exhibits robust performance in most cases.}
    \label{fig:samvisualcrop}
\end{minipage}
\hfill
\begin{minipage}{0.48\textwidth}
    \centering
    \includegraphics[width=\textwidth]{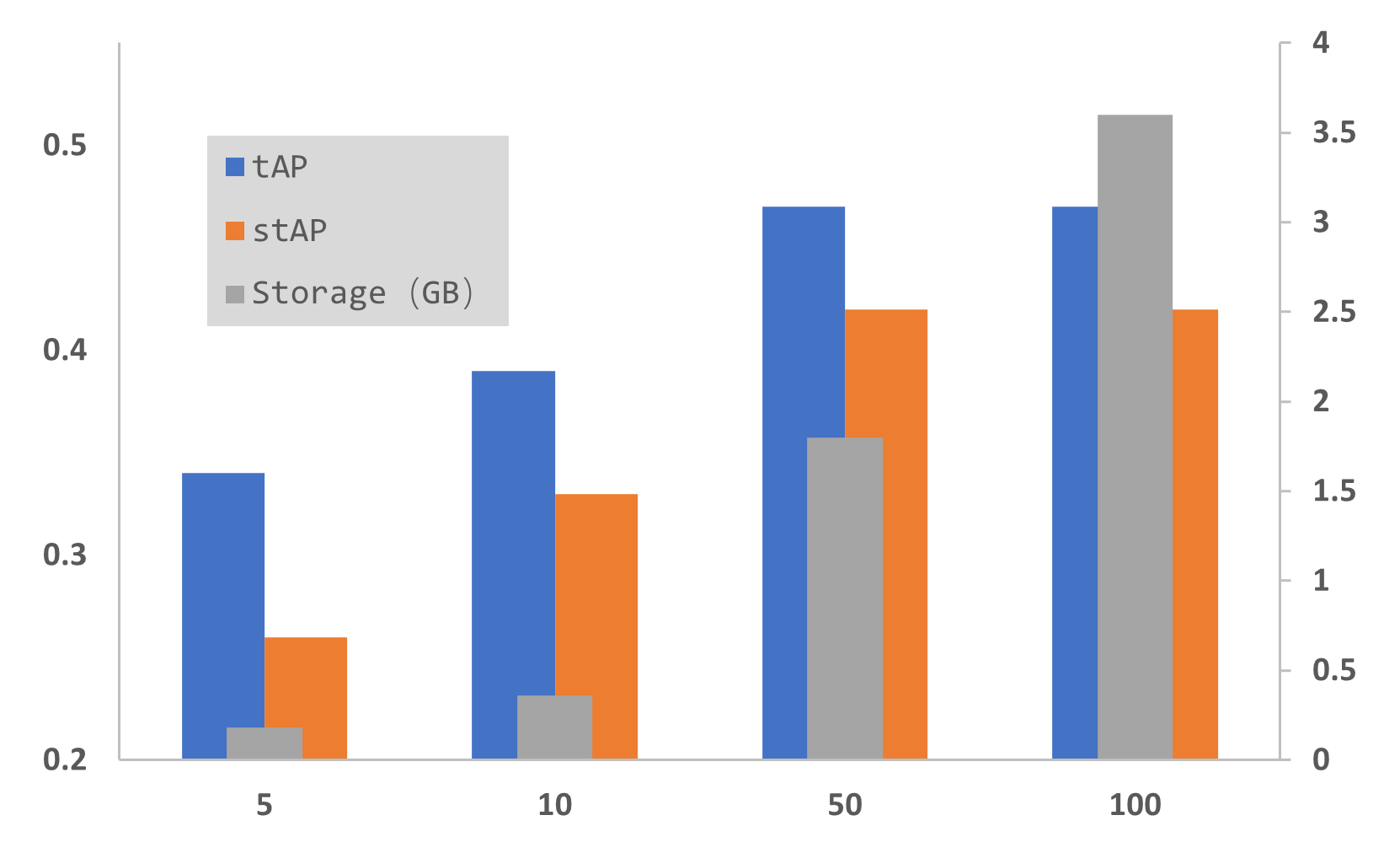}
    \caption{\textbf{Ablation Analysis of Capacity and Storage.} The horizontal axis represents the selected memory bank capacity, the left vertical axis corresponds to the range of VQ2D evaluation metrics, and the right vertical axis indicates storage capacity in gigabytes (GB).}
    \label{fig:capacity}
\end{minipage}
\end{figure}

As shown in Figure.\ref{fig:samvisualcrop}, we employ a combined center-point and bounding box prompting strategy for segmentation. This leverages the assumption that the target object is centrally located in the template, using its center point along with a bounding box derived from a 2/3-scaled visual crop. We also investigated segmentation pathways: the ”everything” prompt, selecting the largest resultant mask; and
”positive/negative point” prompts, using the center and edge
points, respectively.
\begin{figure}[]
\centering
\includegraphics[width=\textwidth,height=0.47\columnwidth]{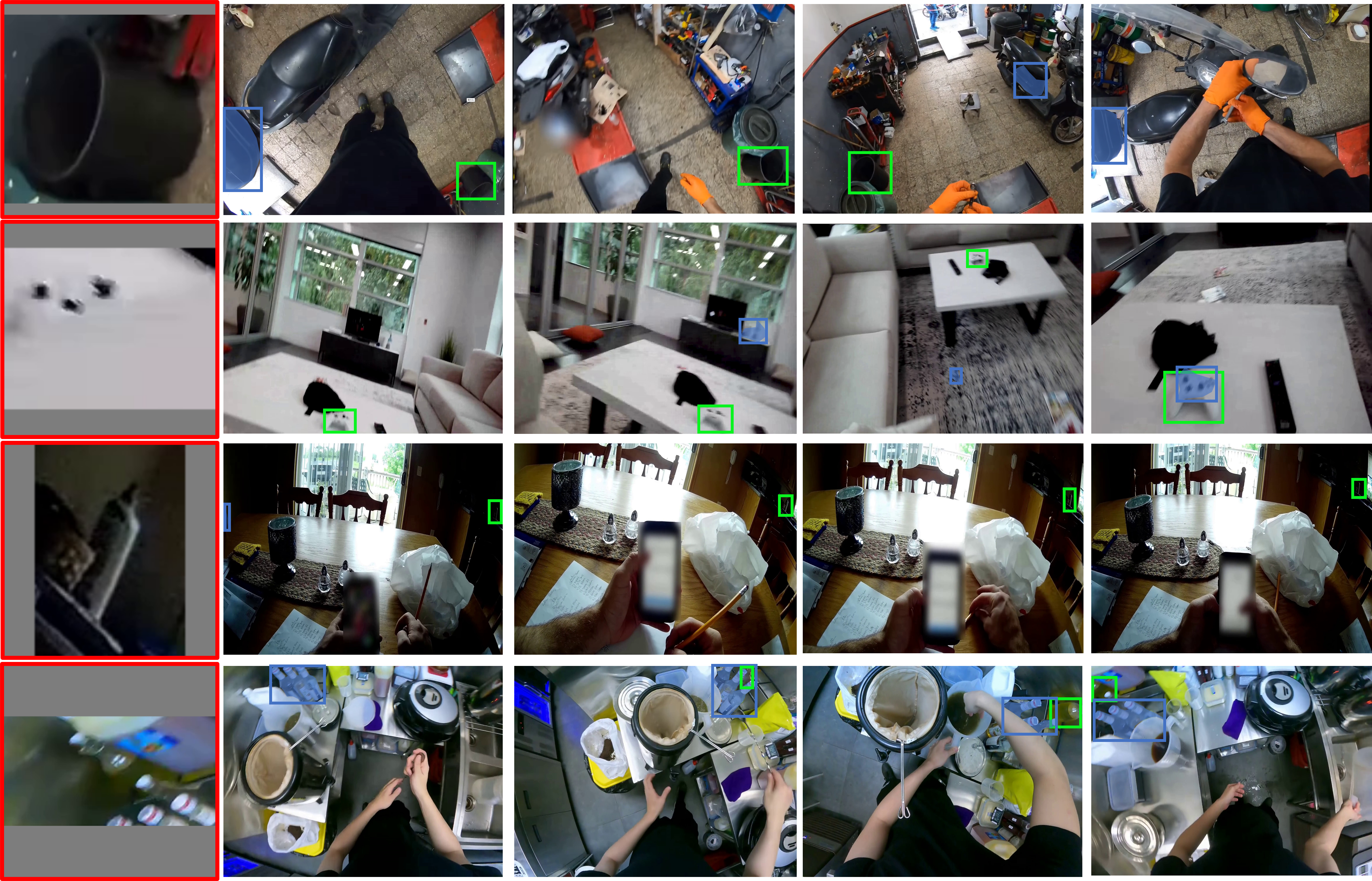}
\caption{\textbf{Visualization of failure examples in Ego4D-VQ2D}.}
\label{fig:failure}
\end{figure}
\subsubsection{Memory bank capacity $O_{MAX}$.}
\label{appendix:capacity}
To facilitate intuitive comparison, we synchronized the maximum memory bank capacities of AMM and GLM, with differences only in their update strategies and content. Figure~\ref{fig:capacity} presents the ablation results on the VQ2D validation set, where we quantified performance by combining accuracy and storage metrics. When the capacity $O$ is small, the memory bank contains only a limited number of “classic appearance” templates, leading to low similarity between new frames and any template when significant changes in pose, illumination, or resolution occur. Conversely, when $O$ is excessively large, the memory bank imposes a heavy storage burden with limited accuracy improvements. We ultimately adopted a capacity of 50.
\subsubsection{Crop ratio settings.}
\label{appendix:cropratio}
From the VQL response trajectory, we first select high-quality segmentation masks whose confidence scores exceed a predefined threshold. Each selected mask is treated as a candidate sample for online updates. To add a candidate to the memory bank, we perform the following cropping and normalization procedure. First, we determine the crop center and size by computing the centroid of the segmentation mask. Using this centroid as the center, we define a square cropping region that encompasses the object's bounding box and includes surrounding context. As shown in Table.\ref{tab:crop}, we evaluate four different scaling ratios: 1.0, 1.44, 2.25, and 4.0 (corresponding to 1.0×, 1.2×, 1.5×, and 2.0× the bounding box side-length, respectively). The results demonstrate that adopting an area scaling factor of 2.25 (i.e., 1.5× side-length scaling) achieves optimal performance gains. The candidate mask is then cropped according to this ratio. If the cropping region extends beyond the original image boundaries, the out-of-bounds area is padded with zeros. To prevent excessive zero-padding from contaminating the memory samples, we enforce a constraint: if the padded area exceeds 50\% of the total crop area, we discard the current scaling factor and iteratively apply the next smaller one from our list (e.g., from 2.25 to 1.44). This process is repeated until the padded area constitutes less than 50\% of the crop or the scaling factor is reduced to 1.0.
\subsubsection{Weighting factor settings}
\label{appendix:rho}
In this section, we investigate the effect of the loss weighting factor $\rho$ on model performance. This experiment, conducted on the VQ2D validation set, aims to find the optimal balance between the segmentation loss $\mathcal{L}_{seg}$ and the tracking loss $\mathcal{L}_{tck}$. As shown in Fig.\ref{fig:rho}, we analyze four weighting configurations: (i) $\rho=0$: The model, trained solely on single-frame recognition, lacks temporal tracking capabilities, leading to poor performance in complex scenarios. (ii) $\rho=0.5$: The introduction of the tracking loss enables the model to learn instance-level features, significantly improving retrieval robustness. (iii) $\rho=1$: The segmentation and tracking losses achieve a synergistic effect. The tracking branch ensures temporal continuity, while the segmentation branch refines the spatial mask boundaries, yielding the best overall performance. (iv) $\rho=2$: The model excels in tracking stability, but the reduced emphasis on segmentation results in coarse mask boundaries, which in turn lowers the Success and stAP metrics that demand high spatial precision. The experiment indicates that $\rho=1$ strikes the optimal balance between temporal consistency and spatial accuracy. Therefore, we adopt this setting for our final model.
\begin{table}[]
\centering
\caption{Ablation study for crop ratio}
\label{tab:crop}
\begin{tabular}{@{}lllll@{}}
\toprule
Ratio & tAP$_{25}\uparrow$  & stAP$_{25}\uparrow$ & Rec.(\%)$\uparrow$ & Succ.(\%)$\uparrow$ \\ \midrule
1.0     & 0.46 & 0.41 & 51.18  & 59.66   \\
1.2   & 0.46 & 0.42 & 51.96  & 61.04   \\
1.5   & 0.47  & 0.42 & 52.09  & 61.29   \\
2.0     & 0.47  & 0.42 & 52.01  & 61.17   \\ \bottomrule
\end{tabular}
\end{table}
\begin{figure}[!h]
\centering
\begin{minipage}{0.48\textwidth}
    \centering
    \includegraphics[width=\textwidth]{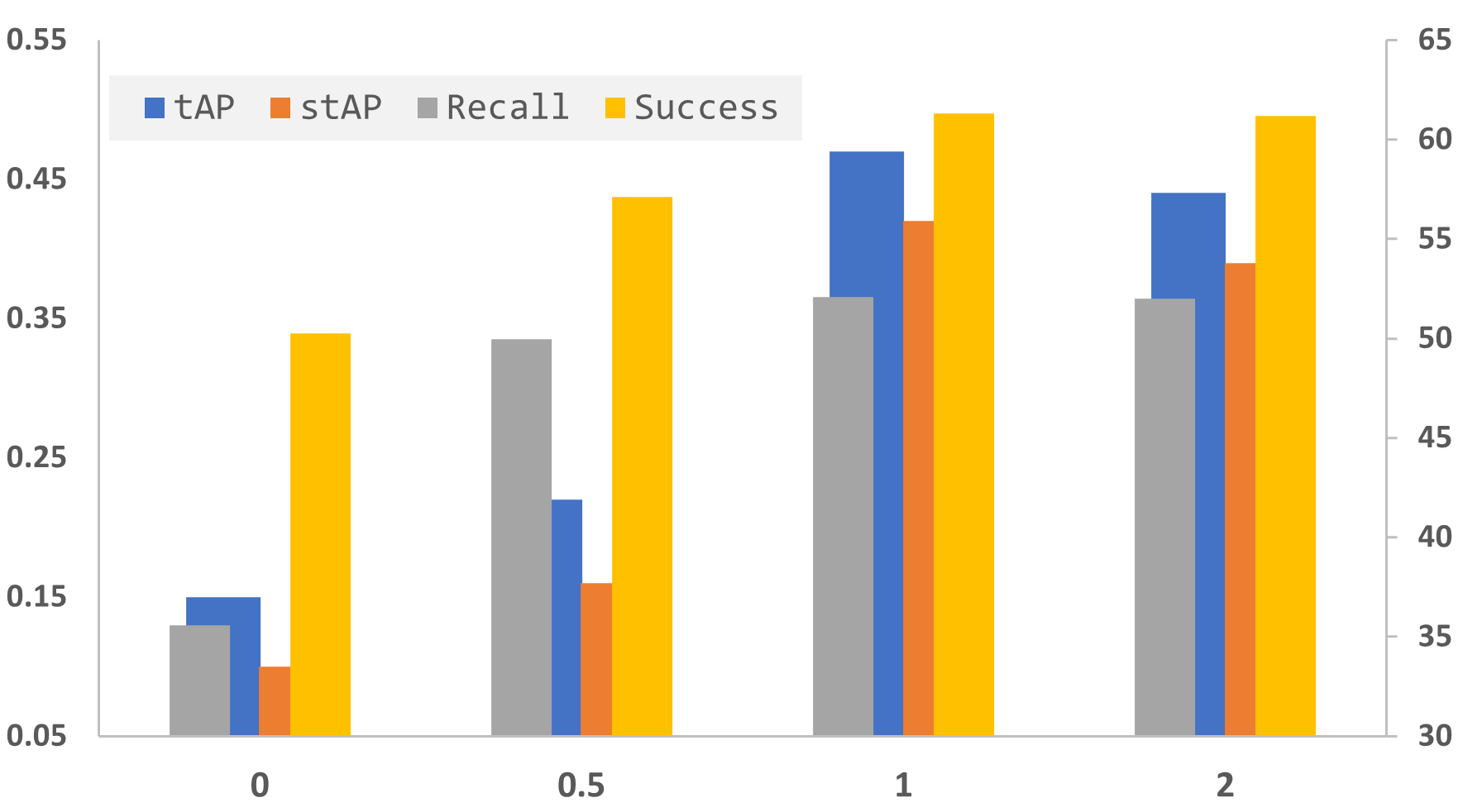}
    \label{fig:rho}
    \caption{\textbf{Ablation study of the weighting factor on the Ego4D-VQ2D validation set}. The left vertical axis denotes the scores for tAP and stAP, while the right vertical axis represents Recall and Success rates in percentage. The horizontal axis corresponds to different settings of the hyperparameter $\rho$.}
    \label{fig:rho}
\end{minipage}
\hfill
\begin{minipage}{0.48\textwidth}
    \centering
    \includegraphics[width=\textwidth]{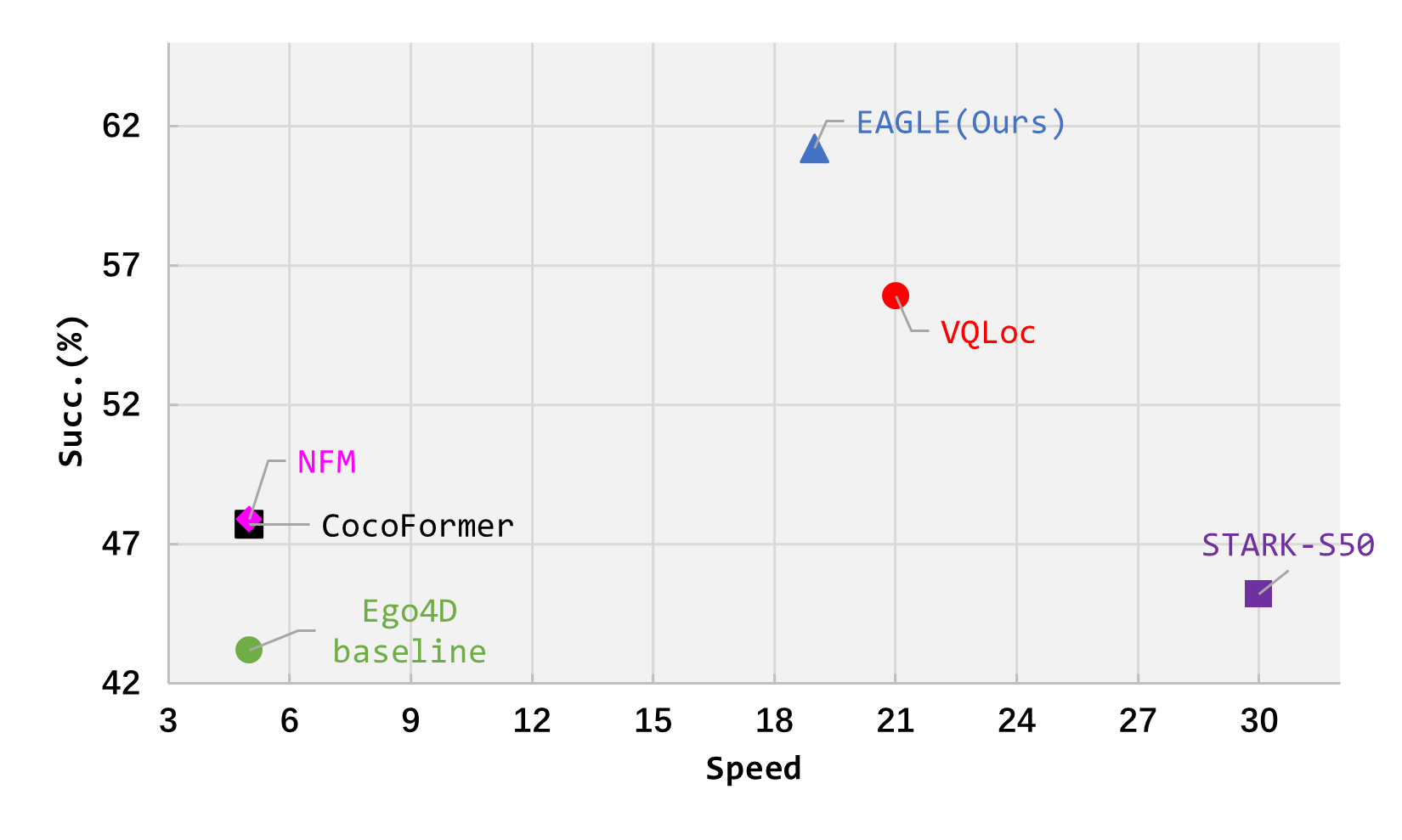}
    \caption{\textbf{Comparison of inference speed on the VQ2D validation set.} The horizontal axis represents runtime speed in FPS, while the vertical axis measures performance using the Success metric from the VQ2D benchmark.}
    \label{fig:speed}
\end{minipage}
\end{figure}
\subsubsection{Inference Speed on Ego4D-VQ}
\label{appendix:speed}
As shown in Figure.\ref{fig:speed}, we compare the inference speed of our method against several state-of-the-art approaches, without employing any acceleration techniques such as TensorRT. Although our model incorporates an online update mechanism, it leverages a highly efficient and fast-converging optimization strategy. This allows the model to acquire sufficient gradient information to guide precise spatio-temporal localization within a few iteration steps and with only a limited number of approximate computations. Consequently, while not achieving strict real-time performance, its runtime efficiency remains comparable to that of other leading methods, demonstrating an effective trade-off between performance and speed.
\subsection{Failure Analysis}
\label{appendix:failure}
Figure.\ref{fig:failure} illustrates several typical failure cases. These failures are predominantly caused by three major challenges inherent in the queries: (1) Intrinsic ambiguity of the query object(e.g., the second row), where the model fails to discern the object's contour and texture, leading it to match visually similar but incorrect candidates across a wide area. (2) Severe illumination changes in the scene (e.g., the first and third rows), which cause textural distortions of the target, resulting in the model either tracking distractors with similar textures or failing the retrieval entirely. (3) Ambiguous or multi-object references in the query (e.g., the fourth row), where the model treats all objects in the template as the target, leading to a trajectory that only partially covers the intended object, thereby severely degrading accuracy. We contend that addressing these challenging scenarios requires the integration of more advanced low-level vision pre-processing techniques in future work, which would substantially enhance the model's robustness.
\end{document}